# Aggregated Wasserstein Metric and State Registration for Hidden Markov Models

Yukun Chen, Jianbo Ye, and Jia Li

**Abstract**—We propose a framework, named *Aggregated Wasserstein*, for computing a dissimilarity measure or distance between two Hidden Markov Models with state conditional distributions being Gaussian. For such HMMs, the marginal distribution at any time position follows a Gaussian mixture distribution, a fact exploited to softly match, aka register, the states in two HMMs. We refer to such HMMs as Gaussian mixture model-HMM (GMM-HMM). The registration of states is inspired by the intrinsic relationship of optimal transport and the Wasserstein metric between distributions. Specifically, the components of the marginal GMMs are matched by solving an optimal transport problem where the cost between components is the Wasserstein metric for Gaussian distributions. The solution of the optimization problem is a fast approximation to the Wasserstein metric between two GMMs. The new Aggregated Wasserstein distance is a semi-metric and can be computed without generating Monte Carlo samples. It is invariant to relabeling or permutation of states. The distance is defined meaningfully even for two HMMs that are estimated from data of different dimensionality, a situation that can arise due to missing variables. This distance quantifies the dissimilarity of GMM-HMMs by measuring both the difference between the two marginal GMMs and that between the two transition matrices. Our new distance is tested on tasks of retrieval, classification, and t-SNE visualization of time series. Experiments on both synthetic and real data have demonstrated its advantages in terms of accuracy as well as efficiency in comparison with existing distances based on the Kullback-Leibler divergence.

**Index Terms**—Hidden Markov Model, Gaussian Mixture Model, Wasserstein Distance, Optimal Transport

## 1 INTRODUCTION

A hidden Markov model (HMM) with Gaussian emission distributions for any given state is a widely used stochastic model for time series of vectors residing in an Euclidean space. It has been widely used in the pattern recognition literature, such as acoustic signal processing (e.g. [1], [2], [3], [4], [5], [6], [7], [8]) and computer vision (e.g. [9], [10], [11], [12]) for modeling spatial-temporal dependencies in data. We refer to such an HMM as Gaussian mixture model-HMM (GMM-HMM) to stress the fact that the marginal distribution of the vector at any time spot follows a Gaussian mixture distribution. Our new distance for HMMs exploits heavily the GMM marginal distribution, which is the major reason we use the terminology GMM-HMM. We are aware that in some literature, Gaussian mixture HMM is used to mean an HMM with state conditional distributions being Gaussian mixtures rather than a single Gaussian distribution. This more general form of HMM is equivalent to an HMM containing an enlarged set of states with single Gaussian distributions. Hence, it poses no particular difficulty for our proposed framework. More detailed remarks are given in Section 6.

A long-pursued question is how to quantitatively compare two sequences based on the parametric representations of the GMM-HMMs estimated from them respectively. The GMM-HMM parameters lie on a non-linear manifold. Thus a simple Euclidean distance on the parameters is not proper. As argued in the literature (e.g. [13], [14]), directly compar-

- *Y. Chen and J. Ye are with the College of Information Sciences and Technology, The Pennsylvania State University.*
  *E-mail:*
- *J. Li is with Department of Statistics, The Pennsylvania State University.*

*Manuscript received April 19, 2017; revised –, 2017.*

ing HMM in terms of the parameters is non-trivial, partly due to the *identifiability* issue of parameters in a mixture model. Specifically, a mixture model can only be estimated up to the permutation of states. Different components in a mixture model are actually unordered even though labels are assigned to them. In other words, the permutation of labels has no effect on the likelihood of the model. Some earlier solutions do not principally tackle the parameter identifiability issue and simply assume the components are already aligned based on whatever labels given to them [15]. Other more sophisticated solutions sidestep the issue to use model independent statistics including the KL divergence [16], [17] and probability product kernels [18], [19]. Those statistics however cannot be computed easily, requiring Monte Carlo samples or the original sequences [14], [20], which essentially serve as Monte Carlo samples.

Sometimes approaches that use the original sequence data give more reliable results than the Monte Carlo approaches. But such approaches require that the original sequences are instantly accessible at the phase of data analysis. Imagine a setting where large volumes of data are collected across different sites. Due to the communication constraints or the sheer size of data, it is possible that one cannot transmit all data to a single site. We may have to work on a distributed platform. The models are estimated at multiple sites; and only the models (much compressed information from the original data) are transmitted to a central site. This raises the need of approaches requiring only the model parameters. Existing methods using only the model parameters typically rely on Monte Carlo sampling (e.g. KL-D based methods [16]) to calculate certain log-likelihood statistics. However, the rate of convergence in estimating the log-likelihoods is $O\left(\left(\frac{1}{n}\right)^{2/d}\right)$ [21], [22], where $n$ is the data size and $d$ the dimension. This can be slow for GMM-HMMs in high dimensions, not to mention the time to generate those samples.

In this paper, we propose a non-simulation parameter-based framework named *Aggregated Wasserstein* to compute the distance between GMM-HMMs. To address the state identifiability issue, the framework first solves a registration matrix between the states of two GMM-HMMs according to an optimization criterion. The optimization problem is essentially a fast approximation to the Wasserstein metric between two marginal GMMs. Once the registration matrix is obtained, we compute separately the difference between the two marginal GMMs and the difference between two transition matrices. Finally, we combine the two parts by a weighted sum. The weight can be cast as a trade-off factor balancing the importance between differentiating spatial geometries and stochastic dynamics of two GMM-HMMs.

Empirical results show that the advantages of the aggregated Wasserstein approach are not restricted to computational efficiency. In fact, the new distance outperforms KL divergence purely as a distance measure under some scenarios. We thus move one step further under this parameter-based framework for defining a distance between HMMs. Aiming at improving how the states are registered, we propose a second approach to calculate the registration matrix based on Monte Carlo sampling. The second approach overcomes certain limitations of the first approach, but at the cost of being more computationally expensive. Despite requiring Monte Carlo samples, the second approach has a rate of convergence asymptotically at $O\left(\sqrt{\frac{\log n}{n}}\right)$ — much faster than the rate of computing log-likelihood based statistics in high dimensions.

We investigate our methods in real world tasks and compare them with the KL divergence-type methods. Practical advantages of our approach have been demonstrated in real applications. By experiments on synthetic data, we also make effort to discover scenarios when our proposed methods outperform the others.

**Our contributions.** We develop a parameter-based framework with the option of not using simulation for computing a distance between GMM-HMMs. Under such framework, a registration matrix is computed for the states in two HMMs. Two methods have been proposed to compute the registration, resulting in two distances, named *Minimized Aggregated Wasserstein* and *Improved Aggregated Wasserstein*. Both distances are experimentally validated to be robust and effective, often outperform KL divergence-based methods in practice.

The rest of the paper is organized as follows. We introduce notations and preliminaries in Section 2. The main framework for defining the distance is proposed in Section 3. The second approach based on Monte Carlo to compute the registration between two sets of HMM states is described in Section 4. Finally, we investigate the new framework empirically in Section 5 based on synthetic and real data.

# 2 PRELIMINARIES

In Section 2.1, we review GMM-HMM and introduce notations. Next, the definition for Wasserstein distance is pro-vided in Section 2.2, and its difference from the KL divergence in the case of Gaussian distributions is discussed.

## 2.1 Notations and Definitions

Consider a sequence $O_T = \{o_1, o_2, ..., o_T\}$ modeled by a GMM-HMM. Suppose there are $M$ states: $S = \{1, ..., M\}$, a GMM-HMM under the stationary condition assumes the following:

1) Each observation $o_i \in O_T$ is associated with a hidden state $s_i \in S$ governed by a Markov chain (MC).

2) $\mathbf{T}$ is the $M \times M$ transition matrix of the MC $\mathbf{T}_{i,j} \overset{\text{def}}{=} P(s_{t+1} = j | s_t = i)$, $1 \leq i, j \leq M$ for any $t \in \{1, ..., T\}$. The stationary (initial) state probability $\pi = [\pi_1, \pi_2, ..., \pi_M]$ satisfies $\pi \mathbf{T} = \pi$ and $\pi \mathbf{1} = 1$.

3) The Gaussian probabilistic emission function $\phi_i(o_t) \overset{\text{def}}{=} P(o_t | s_t = i)$, $i = 1, ..., M$, for any $t \in \{1, ..., T\}$, is the p.d.f. of the normal distribution $\mathcal{N}(\mu_i, \Sigma_i)$, where $\mu_i, \Sigma_i$ are the mean and covariance of the Gaussian distribution conditioned on state $i$.

In particular, we use $\mathcal{M}(\{\mu_i\}_{i=1}^M, \{\Sigma_i\}_{i=1}^M, \pi)$ to denote the corresponding mixture of $M$ Gaussions ( $\{\phi_1, \phi_2, ..., \phi_M\}$ ). As we assume the Markov chain has become stationary, $\mathcal{M}$'s prior probabilities of components, a.k.a. the mixture weights, are determined by $\pi$, the stationary distribution of $\mathbf{T}$. Therefore, one can summarize the parameters for a stationary GMM-HMM model via $\Lambda$ as $\Lambda(\mathbf{T}, \mathcal{M}) = \Lambda(\mathbf{T}, \{\mu_i\}_{i=1}^M, \{\Sigma_i\}_{i=1}^M)$. In addition, the $i$-th row of the transition matrix $\mathbf{T}$ is denoted by $\mathbf{T}(i, :) \in \mathbb{R}^{1 \times M}$. And the next observation's distribution conditioned on current state $i$ is also a GMM: $\mathcal{M}^{(i)}(\{\mu_i\}_{i=1}^M, \{\Sigma_i\}_{i=1}^M, \mathbf{T}(i, :))$, which we abbreviated as $\mathcal{M}^{(i)}|_{\mathbf{T}(i, :)}$.

## 2.2 The Wasserstein Distance and the Gaussian Case

In probability theory, Wasserstein distance is a geometric distance naturally defined for any two probability measures over a metric space.

*Definition 1 (p-Wasserstein distance).* Given two probability distribution $f, g$ defined on Euclidean space $\mathbb{R}^d$, the $p$-Wasserstein distance $W_p(\cdot, \cdot)$ between them is given by

$$W_p(f, g) \overset{\text{def}}{=} \left[ \inf_{\gamma \in \Pi(f,g)} \int_{\mathbb{R}^d \times \mathbb{R}^d} \|\mathbf{x} - \mathbf{y}\|^p d\gamma(\mathbf{x}, \mathbf{y}) \right]^{1/p}, \quad (1)$$

where $\Pi(f, g)$ is the collection of all distributions on $\mathbb{R}^d \times \mathbb{R}^d$ with marginal $f$ and $g$ on the first and second factors respectively. In particular, the $\Pi(\cdot, \cdot)$ is often called as the coupling set. The $\gamma^* \in \Pi(f, g)$ that takes the infimum in Eq. (1) is called the optimal coupling.

Because it takes cross-support relationship into consideration, it has shown strength in computer vision [23], [24], document classification [25], [26], data mining[27] etc.

*Remark 1.* By Hölder inequality, one has $W_p \leq W_q$ for any $p \leq q < \infty$. In this paper, we focus on the practice of $W_p$ with $0 < p \leq 2$. (Please see supplementary material Sec. 1 for a detailed proof of this.)

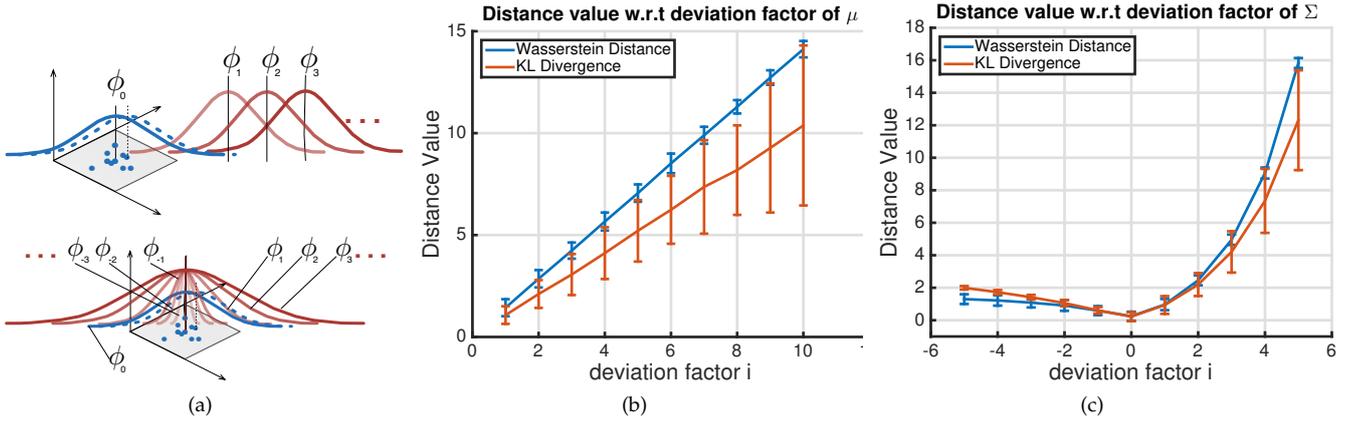

Fig. 1: (a) Experiment scheme for varying $\mu$ and varying $\Sigma$. A re-estimated $\widehat{\phi}_0$ is denoted as the dashed blue line. (b) (c) Mean estimates of $W_2(\widehat{\phi}_0, \phi_i)$ (blue) and $KL(\widehat{\phi}_0, \phi_i)$ (orange) and their $3\sigma$ confidence intervals w.r.t different Gaussian $\phi_i$. (b) is for varying $\mu$, and (c) is for varying $\Sigma$.

While Wasserstein distance between two multi-dimensional GMMs is unsolved, it has a closed formula for two Gaussian $\phi_1(\mu_1, \Sigma_1)$ and $\phi_2(\mu_2, \Sigma_2)$ [28] when $p = 2$:

$$W_2^2(\phi_1, \phi_2) = \|\mu_1 - \mu_2\|^2 + \text{tr}\left[\Sigma_1 + \Sigma_2 - 2\left(\Sigma_1^{\frac{1}{2}}\Sigma_2\Sigma_1^{\frac{1}{2}}\right)^{\frac{1}{2}}\right]. \quad (2)$$

**Remark 2.** The formula of Wasserstein distance between two Gaussians does not involve the inverse-covariance matrix, thus admits the cases of singularity. In comparison, KL divergence between two Gaussian $KL(\phi_1, \phi_2)$ goes to infinity if one of the covariances of $\phi_1$ and $\phi_2$ becomes singular.

**Remark 3.** The Wasserstein distance could also be more statistically robust than KL divergence by comparing the variance of their estimations. To illustrate this point, we conduct two sets of toy experiments.

Our first toy experiment is shown in Fig. 1 (a) upper figure. First, we sample 100 batches of data, each of size 50, from the pre-selected Gaussian $\phi_0 = \mathcal{N}\left([0,0], \begin{pmatrix} 1 & 0 \\ 0 & 1 \end{pmatrix}\right)$. Then, we re-estimate each batch's Gaussian parameters $\widehat{\phi}_0 = \mathcal{N}(\widehat{\mu}, \widehat{\Sigma}) \approx \phi_0$ and calculate $W(\widehat{\phi}_0, \phi_i)$ and $KL(\widehat{\phi}_0, \phi_i)$, in which $\phi_i = \mathcal{N}\left([0.5 \cdot i, 0.5 \cdot i], \begin{pmatrix} 1 & 0 \\ 0 & 1 \end{pmatrix}\right)$, $i = 1, ..., 10$ is a sequence of Gaussians, both with closed forms. Ideally, a distance that can consistently differentiate the $\phi_i$ by computing its distance to the $\widehat{\phi}_0$ should have larger value as $i$ grows. Also, its sample deviations of $W_2(\phi_i, \widehat{\phi}_0)$ or $KL(\phi_i, \widehat{\phi}_0)$ should be small enough to not mask out the change from $i$ to $i + 1$. Fig. 1 (b) shows the performance of Wasserstein Distance and KL divergence on this toy experiment. Both the averaged distance to $\phi_i$ and the $3\sigma$ confidence interval are plotted. It is clear that the Wasserstein distances based on estimated distributions have smaller variance and can overall better differentiate $\{\phi_i\}$.

Likewise, in our second toy experiment, we also conduct a similar toy experiment by changing $\phi_i$'s variances rather than their means (See Fig. 1 (a) bottom figure). At this time, we set $\phi_i = \mathcal{N}\left([0,0], \exp(0.5 \cdot i) \cdot \begin{pmatrix} 1 & 0 \\ 0 & 1 \end{pmatrix}\right)$. The result is plotted in Fig. 1 (c). It shows that KL divergence can be more robust than Wasserstein distance if $\widehat{\phi}_0$ is compared to $\phi_i$ at $i < 0$, but the situation quickly becomes worse at $i \geq 2$. This is due the asymmetric nature of KL divergence. Informally speaking, we conclude from the two toy experiments that estimating $KL(\phi_i, \phi_0)$ can be statistically stable if $\phi_i$ is under the "umbrella" of $\phi_0$, and becomes inaccurate otherwise. On the other hand, Wasserstein distance, as a true metric [29], has consistent accuracy across these two settings.

# 3 THE FRAMEWORK OF AGGREGATED WASSERSTEIN

In this section, we propose a framework to compute the distance between two GMM-HMMs, $\Lambda_1(\mathbf{T}_1, \mathcal{M}_1)$ and $\Lambda_2(\mathbf{T}_2, \mathcal{M}_2)$, where $\mathcal{M}_l$, $l = 1, 2$ are marginal GMMs with pdf $f_l(x) = \sum_{j=1}^{M_l} \pi_{l,j} \phi_{l,j}(x)$ and $\mathbf{T}_1, \mathbf{T}_2$ are the transition matrices of dimension $M_1 \times M_1$ and $M_2 \times M_2$ (recall notations in Section 2 ). Based on the registration matrix between states in two HMMs, to be described in Section 3.1, the distance between $\Lambda_1$ and $\Lambda_2$ consists of two parts: (1) the difference between $\mathcal{M}_1$ and $\mathcal{M}_2$ (Section 3.2); and (2) the difference between $\mathbf{T}_1$ and $\mathbf{T}_2$ (Section 3.3).

## 3.1 The Registration of States

The registration of states is to build a correspondence between $\Lambda_1$'s states and $\Lambda_2$'s states. In the simplest case (an example is illustrated in Fig 2), if the two marginal GMMs are identical distributions but the states are labeled differently (referred to as permutation of states), the registration should discover the permutation and yield a one-one mapping between the states. We can use a matrix $\mathbf{W} = \{w_{i,j}\} \in \mathbb{R}^{M_1 \times M_2}$ whose elements $w_{i,j} \geq 0$ to encode this registration. In particular, $w_{i,j} = \pi_{1,i}(= \pi_{2,j})$ iff state $i$ in $\Lambda_1$ is registered to state $j$ in $\Lambda_2$. With $\mathbf{W}$ given, through matrix multiplications (details delayed in

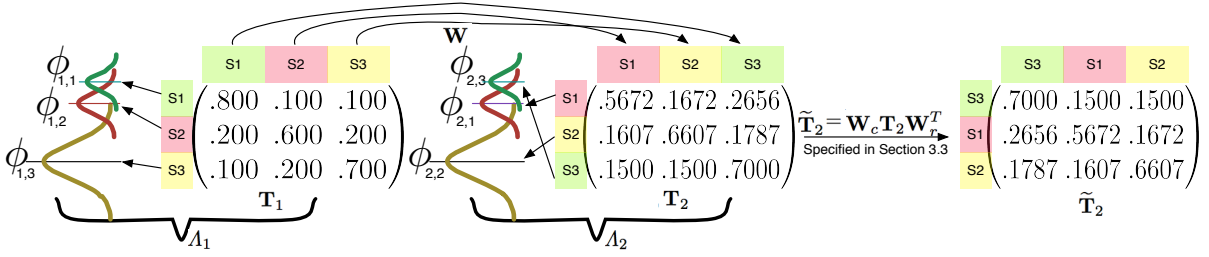

Fig. 2: A simple registration example about how $\mathbf{T}_2$ in $\Lambda_2$ is registered towards $\Lambda_1$ such that it can be compared with $\mathbf{T}_1$ in $\Lambda_1$. For this example, $\mathbf{W}$ encodes a "hard matching" between states in $\Lambda_1$ and $\Lambda_2$

Section 3.3), the rows and columns of $\mathbf{T}_1$ can be permuted to become identical to $\mathbf{T}_2$.

Generally and more commonly, there may exist no state in $\Lambda_2$ having the same emission function as some state in $\Lambda_1$, and the number of states in $\Lambda_1$ may not equal that in $\Lambda_2$. The registration process becomes more difficult. We resort to the principled optimal transport [29] as a tool to solve this problem and formulate the following optimization problem. Recall Eq. (2) for how to compute $W_2(\phi_{1,i}, \phi_{2,j})$. Let $0 < p \leq 2$. Consider

$$\min_{\mathbf{W} \in \Pi(\pi_1, \pi_2)} \sum_{i=1}^{M_1} \sum_{j=1}^{M_2} w_{i,j} W_2^p(\phi_{1,i}, \phi_{2,j}) \qquad (3)$$

where $\Pi(\pi_1, \pi_2) \stackrel{\text{def}}{=}$

$$\Big\{ \mathbf{W} \in \mathbb{R}^{M_1 \times M_2} : \sum_{i=1}^{M_1} w_{i,j} = \pi_{2,j}, j = 1, \ldots, M_2;$$

$$\sum_{j=1}^{M_2} w_{i,j} = \pi_{1,i}, i = 1, \ldots, M_1; \text{ and } \quad w_{i,j} \geq 0, \forall i, j \Big\} \qquad (4)$$

The rationale behind this is that, two states whose emission functions are geometrically close and in similar shape should be more likely to be matched. The solution $\mathbf{W} \in \Pi(\pi_1, \pi_2)$ of the above optimization is called the *registration matrix* between $\Lambda_1$ and $\Lambda_2$. And it will play an important role in the comparison of both marginal GMMs and transition matrices of $\Lambda_1$ and $\Lambda_2$.

The solution of Eq. (3) is an extension of hard matching between states for the simplest case to the general soft matching when the hard matching is impossible. For the aforementioned simple example (Fig. 2), in which the two Gaussian mixtures are in fact identical (thus hard matching is possible), the solution of Eq. (3) is indeed a permutation matrix $\mathbf{W}$ that correctly maps the states in the two models. In general, there are more than one non-zero elements per row or per column.

### 3.2 The Distance between Two Marginal GMMs

Our aim in this subsection is to quantify the difference between $\Lambda_1$ and $\Lambda_2$'s marginal GMMs $\mathcal{M}_1$ and $\mathcal{M}_2$ with density functions $f_1(x) = \sum_{j=1}^{M_1} \pi_{1,j} \phi_{1,j}(x)$ and $f_2(x) = \sum_{j=1}^{M_2} \pi_{2,j} \phi_{2,j}(x)$ respectively.

Given the discussion on the advantages of the Wasserstein metric (especially the Gaussian case) in Section 2, one may ask *why not to use Wasserstein distance* $W(\mathcal{M}_1, \mathcal{M}_2)$

*directly to measure the dissimilarity between* $\mathcal{M}_1, \mathcal{M}_2$? Unfortunately, there is no closed form formula for GMMs except for the reduced case of single Gaussians. Monte Carlo estimation is usually used. However, similar to the estimation of KL divergence, the Monte Carlo estimation for the Wasserstein distance also suffers from a slow convergence rate. The rate of convergence is as slow as that of KL divergence, i.e., $O\left(\left(\frac{1}{n}\right)^{1/d}\right)$ [30], again posing difficulty in high dimensions. So instead of estimating the Wasserstein distance itself, we make use of the solved registration matrix $\mathbf{W} \in \Pi(\pi_1, \pi_2)$ (from Eq. (3)) and the closed form Wasserstein distance between every pair of Gaussians to quantify the dissimilarity between two marginal GMMs $\mathcal{M}_1$ and $\mathcal{M}_2$. The rationale is that the matching weights in $\mathbf{W}$ establish a correspondence between the components in the two GMMs, under which a straightforward summation (of course, proper normalization is guaranteed) of the pairwise distances between the matched Gaussian components quantifies the dissimilarity between the GMMs. Specifically, we define the *registered distance* between $\mathcal{M}_1$ and $\mathcal{M}_2$ at $\mathbf{W}$:

$$\widetilde{R}_p^p(\mathcal{M}_1, \mathcal{M}_2; \mathbf{W}) \stackrel{\text{def}}{=} \sum_{i=1}^{M_1} \sum_{j=1}^{M_2} w_{i,j} W_2^p(\phi_{1,i}, \phi_{2,j}) \qquad (5)$$

Note that registration matrix solved by a scheme other than Eq. (3) (e.g. the one we will introduce in Section 4) can also be plugged into this equation. We will later prove that $\widetilde{R}_p(\mathcal{M}_1, \mathcal{M}_2; \mathbf{W})$ is a semi-metric (Theorem 2). Next, we present Theorem 1 that states that this semi-metric is an upper bound on the true Wasserstein metric.

*Theorem 1.* For any two GMMs $\mathcal{M}_1$ and $\mathcal{M}_2$, define $\widetilde{R}_p(\cdot, \cdot : \mathbf{W})$ by Eq. (5). If $\mathbf{W} \in \Pi(\pi_1, \pi_2)$, we have for $0 < p \leq 2$

$$\widetilde{R}_p(\mathcal{M}_1, \mathcal{M}_2 : \mathbf{W}) \geq W_p(\mathcal{M}_1, \mathcal{M}_2),$$

where $W_p(\mathcal{M}_1, \mathcal{M}_2)$ is the true Wasserstein distance between $\mathcal{M}_1$ and $\mathcal{M}_2$ as defined in Eq. (1).

*Proof 1.* By Remark 1, We have:

$$\widetilde{R}_p^p(\mathcal{M}_1, \mathcal{M}_2; \mathbf{W}) \geq \sum_{i=1}^{M_1} \sum_{j=1}^{M_2} w_{i,j} W_p^p(\phi_{1,i}, \phi_{2,j}) \qquad (6)$$

We construct $\gamma \in \Pi(\mathcal{M}_1, \mathcal{M}_2)$ in the following way: Given a $\mathbf{W} \in \Pi(\pi_1, \pi_2)$ and any $\gamma_{i,j} \in \Pi(\phi_{1,i}, \phi_{2,j})$ for

$i = 1, \ldots, M_1$ and $j = 1, \ldots, M_2$, we let $\widetilde{\Pi}(\mathcal{M}_1, \mathcal{M}_2) =$

$$\left\{ \gamma \overset{\text{def}}{=} \sum_{i=1}^{M_1} \sum_{j=1}^{M_2} w_{i,j} \gamma_{i,j} \,\middle|\, \mathbf{W} \in \Pi(\pi_1, \pi_2), \text{ and} \right.$$

$$\left. \gamma_{i,j} \in \Pi(\phi_{1,i}, \phi_{2,j}),\, i = 1, \ldots, M_1,\, j = 1, \ldots, M_2 \right\} \quad (7)$$

and $\sum_{i=1}^{M_1} \sum_{j=1}^{M_2} w_{i,j} W_p^p(\phi_{1,i}, \phi_{2,j})$ is the exact infimum for all possible $\gamma \in \widetilde{\Pi}(\mathcal{M}_1, \mathcal{M}_2)$, where we see $\widetilde{\Pi}(\mathcal{M}_1, \mathcal{M}_2) \subseteq \Pi(\mathcal{M}_1, \mathcal{M}_2)$. Thus,

$$\sum_{i=1}^{M_1} \sum_{j=1}^{M_2} w_{i,j} W_p^p(\phi_{1,i}, \phi_{2,j}) \geq W_p^p(\mathcal{M}_1, \mathcal{M}_2) \quad (8)$$

By combining it with Eq. (6), the inequality is implied. ∎

For the brevity of notation, if $\mathbf{W}$ is solved from Eq. (3), the resulting distance $\widetilde{R}_p(\mathcal{M}_1, \mathcal{M}_2; \mathbf{W})$ is denoted by $\widetilde{W}_p(\mathcal{M}_1, \mathcal{M}_2)$.

### 3.3 The Distance between Two Transition Matrices

Given the registration matrix $\mathbf{W}$, our aim in this subsection is to quantify the difference between $\Lambda_1$ and $\Lambda_2$'s transition matrices, $\mathbf{T}_1 \in \mathbb{R}^{M_1 \times M_1}$ and $\mathbf{T}_2 \in \mathbb{R}^{M_2 \times M_2}$. Since the *identifiability* issue is already addressed by the registration matrix $\mathbf{W}$, $\mathbf{T}_2$ can now be registered towards $\mathbf{T}_1$ by the following transform:

$$\widetilde{\mathbf{T}}_2 \overset{\text{def}}{=} \mathbf{W}_r \mathbf{T}_2 \mathbf{W}_c^T \in \mathbb{R}^{M_1 \times M_1}, \quad (9)$$

where matrix $\mathbf{W}_r$ and $\mathbf{W}_c$ are row-wise and column-wise normalized $\mathbf{W}$ respectively, a.k.a. $\mathbf{W}_r = \text{diag}^{-1}(\mathbf{W} \cdot \mathbf{1}) \cdot \mathbf{W}$ and $\mathbf{W}_c = \mathbf{W} \cdot \text{diag}^{-1}(\mathbf{1}^T \cdot \mathbf{W})$. A simple example of this process is illustrated in the right part of Fig. 2. Likewise, $\mathbf{T}_1$ can also be registered towards $\mathbf{T}_2$:

$$\widetilde{\mathbf{T}}_1 \overset{\text{def}}{=} \mathbf{W}_c^T \mathbf{T}_1 \mathbf{W}_r \in \mathbb{R}^{M_2 \times M_2}. \quad (10)$$

Then, a discrepancy denoted by $D(\mathbf{T}_1, \mathbf{T}_2; \mathbf{W})$ to measure the dissimilarity of two transition matrices is defined:

$$D_p^p(\mathbf{T}_1, \mathbf{T}_2 : \mathbf{W}) \overset{\text{def}}{=} d_T^p(\mathbf{T}_1, \widetilde{\mathbf{T}}_2) + d_T^p(\mathbf{T}_2, \widetilde{\mathbf{T}}_1) \quad (11)$$

where $\widetilde{\mathbf{T}}_1$ and $\widetilde{\mathbf{T}}_2$ are calculated from Eq. (9) and Eq. (10) (with $\mathbf{W}$ given) respectively and

$$d_T^p(\mathbf{T}_1, \widetilde{\mathbf{T}}_2) \overset{\text{def}}{=} \sum_{i=1}^{M_1} \pi_{1,i} \widetilde{W}_p \left( \mathcal{M}_1^{(i)} |_{\mathbf{T}_1(i,:)}, \mathcal{M}_1^{(i)} |_{\widetilde{\mathbf{T}}_2(i,:)} \right) \quad (12)$$

$$d_T^p(\mathbf{T}_2, \widetilde{\mathbf{T}}_1) \overset{\text{def}}{=} \sum_{i=1}^{M_2} \pi_{2,i} \widetilde{W}_p \left( \mathcal{M}_2^{(i)} |_{\mathbf{T}_2(i,:)}, \mathcal{M}_2^{(i)} |_{\widetilde{\mathbf{T}}_1(i,:)} \right) \quad (13)$$

We remind that by the notations in Section 2.1, $\mathcal{M}_1^{(i)} |_{\mathbf{T}_1(i,:)}$ is the pdf of the next observation conditioned on the previous state being $i$ (likewise for the other similar terms).

**Remark 4.** One might have noticed that the distances $d_T(\mathbf{T}_1, \widetilde{\mathbf{T}}_2)$ and $d_T(\mathbf{T}_2, \widetilde{\mathbf{T}}_1)$, as defined in Eq. (12) and (13), are not determined totally by the transition matrices after registration has been applied. Take $d_T^p(\mathbf{T}_1, \widetilde{\mathbf{T}}_2)$ as an example. It is a weighted sum over $\widetilde{W}_p^p \left( \mathcal{M}_1^{(i)} |_{\mathbf{T}_1(i,:)}, \mathcal{M}_1^{(i)} |_{\widetilde{\mathbf{T}}_2(i,:)} \right)$, which depends on both the $i$th rows of $\mathbf{T}_1$ and $\widetilde{\mathbf{T}}_2$ and the Gaussian components. Apparent simpler alternatives to $\widetilde{W}_p^p \left( \mathcal{M}_1^{(i)} |_{\mathbf{T}_1(i,:)}, \mathcal{M}_1^{(i)} |_{\widetilde{\mathbf{T}}_2(i,:)} \right)$ include $L_p$ distances or KL divergence between the rows of $\mathbf{T}_1$ and $\widetilde{\mathbf{T}}_2$. The motivation for our more sophisticated treatment is that the states in HMMs often have no actual physical meaning and can suffer from artifacts of parameterization. To measure the similarity between two states, it is more robust to compare the conditional distributions of the observations at the next time position given the states (specified by $\mathcal{M}_1^{(i)} |_{\mathbf{T}_1(i,:)}$ and $\mathcal{M}_1^{(i)} |_{\widetilde{\mathbf{T}}_2(i,:)}$) than the conditional distributions of the next states (specified by the rows of $\mathbf{T}_1$ and $\widetilde{\mathbf{T}}_2$).

### 3.4 A Semi-metric between GMM-HMMs — Minimized Aggregated Wasserstein (MAW)

In summary, the dissimilarity between GMM-HMMs $\Lambda_1, \Lambda_2$ comprises two parts: the first is the discrepancy between the marginal GMMs $\mathcal{M}_1, \mathcal{M}_2$, and the second is the discrepancy between two transition matrices after state registration. A weighted sum of these two terms is taken as the final distance. We call this new distance the *Minimized Aggregated Wasserstein* (MAW) between GMM-HMM models. Let $\mathbf{W}$ be solved from Eq. (3).

$$\begin{aligned} \text{MAW}(\Lambda_1, \Lambda_2) \overset{\text{def}}{=} &(1 - \alpha) \widetilde{R}_p(\mathcal{M}_1, \mathcal{M}_2; \mathbf{W}) \\ &+ \alpha D_p(\mathbf{T}_1, \mathbf{T}_2; \mathbf{W}). \end{aligned} \quad (14)$$

Theorem 2 states that MAW is a semi-metric. A semi-metric shares all the properties of a true metric ( including separation axiom) except for the triangle inequality.

**Theorem 2.** MAW defined by Eq. (14) is a semi-metric for GMM-HMMs if $0 < \alpha < 1$.

**Proof 2.** See Appendix A.

For clarity, we summarize MAW's computation procedure in Algorithm 1.

---

**Algorithm 1** Minimized Aggregated Wasserstein (MAW)

---

**Input:** Two HMMs $\Lambda_1 \left( \mathbf{T}_1, \mathcal{M}_1 \left( \{\mu_{1,i}\}_{i=1}^{M_1}, \{\Sigma_{1,i}\}_{i=1}^{M_1} \right) \right)$ and $\Lambda_2 \left( \mathbf{T}_2, \mathcal{M}_2 \left( \{\mu_{2,i}\}_{i=1}^{M_2}, \{\Sigma_{2,i}\}_{i=1}^{M_2} \right) \right)$
**Output:** $\text{MAW}(\Lambda_1, \Lambda_2) \in \{0\} \cup \mathbb{R}^+$
  1: Compute registration matrix $\mathbf{W}$ by Eq. (3)
  2: Compute $\widetilde{R}_p(\mathcal{M}_1, \mathcal{M}_2; \mathbf{W})$ by Eq. (5)
  3: Compute $D_p(\mathbf{T}_1, \mathbf{T}_2)$ by Eq. (11), Eq. (12) and Eq. (13)
  4: Compute and return $\text{MAW}(\Lambda_1, \Lambda_2)$ defined by Eq. (14).

---

#### 3.4.1 Choosing $\alpha$

The choice of $\alpha$ can depend on the problem in consideration. In the context of classification, we aim at the best capability of distinguishing the classes. In particular, $\alpha$ can be determined by maximizing the classification accuracy obtained by the 1-nearest neighbor classifier on a set of small but representative training GMM-HMMs with ground truth labels. More complex options such as maximizing the area under the receiver operating characteristics curve (AUC) can also be used. For simplicity, we choose $\alpha$ by maximizing 1-nearest neighbor accuracy in this paper. An example for choosing $\alpha$ in our Motion Retrieval experiment is shown in the supplementary material Sec. 2.

## 4 IMPROVED STATE REGISTRATION

The main disadvantage of estimating the matching matrix $W$ by Eq. (3) and then computing $\widetilde{R}_p(\cdot, \cdot)$ by Eq. (5) is that $W$ can be sensitive to the parametrization of GMMs. Two GMMs whose distributions are close can be parameterized very differently, especially when the components are not well separated, resulting in a substantially larger value of $\widetilde{W}$ than the true Wasserstein metric. In contrast, the real Wasserstein metric $W$ only depends on the underlying distributions $\mathcal{M}_{1,2}$, and thus does not suffer from the artifacts caused by the GMM parameterization.

So a key question is "*Can we propose a meaningful interpretation of component-wise matching such that the optimal coupling from $\Pi(\mathcal{M}_1, \mathcal{M}_2)$ for the Wasserstein distance between the two mixture distributions can be realized or approximated arbitrarily well by simulation?*" Interestingly, the answer is yes, but we must revise the notion of matching between two individual components. The resulting new approach is called *Improved Aggregated Wasserstein* (IAW), which approximates the true Wasserstein metric by Monte Carlo simulation.

In the proof of Theorem 1, we see that in the construction of $\widetilde{\Pi}$ (Eq. (7)) we introduced coupling $\gamma_{i,j} \in \Pi(\phi_{1,i}, \phi_{2,j})$ for all pairs of components $\phi_{1,i}$ and $\phi_{2,j}$. The overall coupling between the two distributions is

$$\gamma = \sum_{i=1}^{M_1} \sum_{j=1}^{M_2} w_{i,j} \gamma_{i,j}, \tag{15}$$

which decomposes coupling $\gamma$ into two stages. The first stage contains a set of couplings $\gamma_{i,j}$ between component densities $\phi_{1,i}$ and $\phi_{2,j}$ (matching at the level of data points), while the second stage is the component-wise (or component-level) matching specified by $W$. To maintain a proper meaning of the registration matrix $W$, we would like to keep such a two stage decomposition for $\gamma$. However, when MAW distance is defined, we have a rigid requirement on $\gamma_{i,j}$: $\gamma_{i,j} \in \Pi(\phi_{1,i}, \phi_{2,j})$, which is at the cost of not being able to approach the true Wasserstein metric. Here, we relax this constraint to the extent that the component densities $\phi_{1,i}$ and $\phi_{2,i}$ are still respected.

**Definition 2.** Given $W \in \Pi(\pi_i, \pi_j)$, we say densities $\{\tilde{\phi}_{1,i,j}, \tilde{\phi}_{2,i,j}\}$ couple with $W$ subject to $(\mathcal{M}_1, \mathcal{M}_2)$ if for all $i = 1, \dots, M_1$ and $j = 1, \dots, M_2$,

$$\sum_{j=1}^{M_2} \frac{w_{i,j}}{\pi_{1,i}} \tilde{\phi}_{1,i,j} = \phi_{1,i}, \quad \sum_{i=1}^{M_1} \frac{w_{i,j}}{\pi_{2,j}} \tilde{\phi}_{2,i,j} = \phi_{2,j}. \tag{16}$$

We denote these conditions collectively by

$$\{\tilde{\phi}_{1,i,j}, \tilde{\phi}_{2,i,j}\} \in \Gamma(W \in \Pi(\pi_i, \pi_j) | \mathcal{M}_1, \mathcal{M}_2). \tag{17}$$

Now when we match component $i$ with $j$, instead of treating $\phi_{1,i}$ and $\phi_{2,j}$, we treat $\tilde{\phi}_{1,i,j}$ and $\tilde{\phi}_{2,i,j}$. Figuratively speaking, we divide $\phi_{1,i}$ into $M_2$ parts $\tilde{\phi}_{1,i,j}, j = 1, \dots, M_2$, and $\phi_{2,j}$ into $M_1$ parts $\tilde{\phi}_{2,i,j}, i = 1, \dots, M_1$ subject to the constraints in Eqs. (16) and then match $\tilde{\phi}_{1,i,j}$ and $\tilde{\phi}_{2,i,j}$. Obviously, MAW is a special case where we have identical $\tilde{\phi}_{1,i,j}$'s over all the $j$'s: $\tilde{\phi}_{1,i,j} \equiv \phi_{1,i}$ and similarly $\tilde{\phi}_{2,i,j} \equiv \phi_{2,j}$ for all the $i$'s. This also shows that $\Gamma(W \in \Pi(\pi_i, \pi_j) | \mathcal{M}_1, \mathcal{M}_2)$ is nonempty for any $W \in \Pi(\pi_i, \pi_j)$.

Let $\gamma_{i,j}$ be any coupling measure from $\Pi(\tilde{\phi}_{1,i,j}, \tilde{\phi}_{2,i,j})$, that is, $\tilde{\phi}_{1,i,j}$ and $\tilde{\phi}_{2,i,j}$ are marginals of $\gamma_{i,j}$. If $\{\tilde{\phi}_{1,i,j}, \tilde{\phi}_{2,i,j}\}$ couple with $W$ subject to $(\mathcal{M}_1, \mathcal{M}_2)$, we immediately have

$$\gamma = \sum_{i=1}^{M_1} \sum_{j=2}^{M_2} w_{i,j} \gamma_{i,j} \in \Pi(\mathcal{M}_1, \mathcal{M}_2).$$

Alternatively, instead of defining $\gamma_{i,j} \in \Pi(\tilde{\phi}_{1,i,j}, \tilde{\phi}_{2,i,j})$, we can bypass the introduction of $\tilde{\phi}_{1,i,j}$ and $\tilde{\phi}_{2,i,j}$ and simply impose the following constraints:

$$\sum_{i=1}^{M_1} w_{i,j} \int_{\mathbf{x}} d\gamma_{i,j}(\mathbf{x}, \mathbf{y}) = \pi_{2,j} \phi_{2,j}(\mathbf{y}), \tag{18}$$

$$\sum_{j=1}^{M_2} w_{i,j} \int_{\mathbf{y}} d\gamma_{i,j}(\mathbf{x}, \mathbf{y}) = \pi_{1,i} \phi_{1,i}(\mathbf{x}). \tag{19}$$

In fact, any $\gamma \in \Pi(\mathcal{M}_1, \mathcal{M}_2)$ can be represented in the form of Eq. (15) with $\gamma_{i,j}$ satisfying Eqs. (18) and (19), and the corresponding $W$ is given by

$$W = \int \pi(\mathbf{x}; \mathcal{M}_1) \cdot \pi(\mathbf{y}; \mathcal{M}_2)^T d\gamma(\mathbf{x}, \mathbf{y}), \tag{20}$$

where $\pi(\mathbf{x}; \cdot)$ (a column vector) denotes the posterior mixture component probabilities at point $\mathbf{x}$ inferred from a given GMM.

We now state the following theorem which we will prove in Appendix B.

**Theorem 3.** For any $\gamma \in \Pi(\mathcal{M}_1, \mathcal{M}_2)$, let $W$ be defined by Eq. (20), then there exist $\gamma_{i,j}, i = 1, \dots, M_1$ and $j = 1, \dots, M_2$ satisfying constraints in Eq. (18) and (19) such that $\gamma = \sum_{i=1}^{M_1} \sum_{j=1}^{M_2} w_{i,j} \gamma_{i,j}$.

Suppose the Wasserstein distance between two GMMs $\mathcal{M}_1$ and $\mathcal{M}_2$ are pre-solved such that the inference for their optimal coupling $\gamma^*$ (referring to Definition 1) is at hand. We denote the new state registration matrix induced from $\gamma^*$ by Eq. (20) as $W^*$.

A Monte Carlo method to estimate $W^*$ is hereby given. Two sets ($\{\mathbf{x}_1, \dots, \mathbf{x}_n\}$ and $\{\mathbf{y}_1, \dots, \mathbf{y}_n\}$) of equal size i.i.d. samples are generated from $\mathcal{M}_1$ and $\mathcal{M}_2$ respectively. The $W^*$ is then empirically estimated by

$$\widetilde{W}_n^* \stackrel{\text{def}}{=} [\pi(\mathbf{x}_1; \mathcal{M}_1), \dots, \pi(\mathbf{x}_n; \mathcal{M}_1)] \cdot \Pi_n \\ \cdot [\pi(\mathbf{y}_1; \mathcal{M}_2), \dots, \pi(\mathbf{y}_n; \mathcal{M}_2)]^T \tag{21}$$

where $\Pi_n \in \mathbb{R}^{n \times n}$ is the $p$-th optimal coupling solved for the two samples (essentially a permutation matrix). We use Sinkhorn algorithm to approximately solve the optimal coupling [31]. $\widetilde{W}_n^*$ converges to $W^*$ with probability 1, as $n \to \infty$. Consequently, IAW is defined similarly as MAW in Eq. (14) but with a different $W$ computed from Eq. (21).

**Remark 5 (Convergence Rate).** The estimation of $W^*$ follows the mixture proportion estimation setting [32], [33], whose rate of convergence is $O\left(\sqrt{\frac{V_{\widetilde{\Pi}} \log n}{n}}\right)$. Here $V_{\widehat{\Pi}} = V_{\widehat{\Pi}}(d, M_1, M_2)$ is the VC dimension of the geometric class induced by the family $\widehat{\Pi}(\mathcal{M}_1, \mathcal{M}_2)$ (See Appendix B and [34] for related definitions).

## 5 EXPERIMENTS

We conduct experiments to quantitatively evaluate the proposed MAW and IAW. In particular, we set $p = 1$. Our comparison baseline is KL based distance [16] since it is the most widely used one (e.g. [14],[13]). In Section 5.1, we use synthetic data to evaluate the sensitivity of MAW and IAW to the perturbation of $\mu$, $\Sigma$, and $\mathbf{T}$. Similar synthetic experiments have been done in related work (e.g. [13]). In Section 5.2, we present results of more extensive experiments to confirm the robustness of the findings in Section 5.1 when HMMs are of different numbers of states, dimensions and levels of difficulty to differentiate. In Section 5.3, we compare MAW and IAW with KL using the Mocap data under both retrieval and classification settings. In Section 5.4, we compare MAW and IAW with KL using TIMIT speech data under the settings of t-SNE visualization and $k$ nearest neighbor classification. Note that for KL and IAW, for both of which the sampling size has to be determined, we make sure the sample size is large enough such that the value of distance value has converged.

### 5.1 Sensitivity to the Perturbation of HMM Parameters.

Three sets of experiments are conducted to evaluate MAW and IAW's sensitivity to the perturbation of GMM-HMM parameters $\{\mu_j\}_{j=1}^M$, $\{\Sigma\}_{j=1}^M$, and $\mathbf{T}$ respectively. In each set of experiments, we have five pre-defined 2-state GMM-HMM models $\left\{ \Lambda_i \left( \{\mu_{i,j}\}_{j=1}^2, \{\Sigma_{i,j}\}_{j=1}^2, \mathbf{T}_i \right) \right\}_{i=1}^5$, among which the only difference is GMM means $\{\mu_{i,1}, \mu_{i,2}\}$, covariances $\{\Sigma_{i,1}, \Sigma_{i,2}\}$, or transition matrices $\mathbf{T}_i$. For example, in the first experiment, we perturb $\{\mu_{i,1}, \mu_{i,2}\}$ by setting the 5 GMM-HMM's $\{\mu_{i,1}, \mu_{i,2}\}_{i=1}^5$ to be

$$\left\{ \left\{ \begin{pmatrix} 2 + i\Delta\mu \\ 2 + i\Delta\mu \end{pmatrix}, \begin{pmatrix} 5 + i\Delta\mu \\ 5 + i\Delta\mu \end{pmatrix} \right\} | i = 1, 2, 3, 4, 5 \right\} \quad (22)$$

respectively. $\{\Sigma_{i,1}, \Sigma_{i,2}\}$ for $i = 1, ..., 5$ are the same: $\left\{ \begin{pmatrix} 1 & 0 \\ 0 & 1 \end{pmatrix}, \begin{pmatrix} 1 & 0 \\ 0 & 1 \end{pmatrix} \right\}$. And the transition matrices are also the same: $\begin{pmatrix} 0.8 & 0.2 \\ 0.2 & 0.8 \end{pmatrix}$. $\Delta\mu$ here is a parameter to control the difference between the 5 models. The smaller the value, the more similar are the 5 models and the more challenging will retrieval be. We choose $\Delta\mu$ to be 0.2, 0.4, 0.6, and compare KL, MAW and IAW under each setting. Please refer to Table 1 for detailed experiment setup for the other two experiments. For each of the five models, 10 sequences of dimension 2 and of length 100 are generated. The models estimated from the 10 sequences by the well-known Baum-Welch algorithm become instances that belong to one class. To summarize, we have 50 estimated models in total which belong to 5 classes. Then we use every model as a query to retrieve other models using distances based on KL, MAW and IAW respectively. A retrieved model is considered a match if it is in the same class as the query model (this applies to all the retrieval experiments in the sequel). The precision recall plot for the retrieval are shown in Fig. 3.

As shown by the first row of Fig. 3, IAW performs slightly better than KL, and KL performs better than MAW in the task of differentiating HMMs under the perturbation of $\{\mu_j\}_{j=1}^M$. The third row of Fig. 3 shows that MAW and IAW perform better than KL to differentiate the perturbation of $\mathbf{T}$. From the second row, we see that for the task of differentiating $\{\Sigma\}_{j=1}^M$, KL performs better than IAW, and both IAW and KL perform much better than MAW. But for the less challenging cases, IAW has comparable performance with KL.

The computation time for each algorithm is provided in Table 2. MAW and IAW are implemented in MATLAB and KL divergence is implemented in C.

### 5.2 Robustness with Respect to Dimension, Number of States, and Level of Differentiation

To verify if the observations from the perturbation experiments in Section 5.1 hold for HMMs of different numbers of states and of different dimensions, we conduct more retrieval experiments similar to Section.5.1. These experiments are under different settings of 1) number of states, 2) dimension, 3) level of differentiation between HMM classes/categories.

The HMMs used for retrieval are still drawn from a two-stage scheme similar to that described in Section 5.1. In the first stage, however, we design procedures to automatically generate seed HMMs instead of manually setting them as in Section 5.1. In the second stage, for each seed HMM, we generate $m$ sequences $o_{1:T}$ of length $T$ and estimate an HMM for each sequence. Those estimated HMMs will deviate somewhat from the original seed HMMs, and the amount of deviation can be controlled by $T$ (roughly speaking, the larger the $T$, the less the deviation). Again, each seed HMM is treated as one class. Specifically, the models estimated from each of the $m$ sequences generated by this seed HMM are instances in one class.

The difficulty of retrieval is determined by both the difference between seed HMMs and the variation among the estimated HMMs within one class. The two stage scheme to generate HMMs allows us to control the process of generating a collection of HMMs for retrieval and to facilitate experiments in a larger space of hyper-parameters. For each collection of HMMs generated, we conduct 1-nearest neighbor retrieval as we have done in Section 5.1 to test the differentiation ability of KL, MAW and IAW.

The procedure for generating seed HMMs for $\mu$ perturbation experiments are specified in Algorithm 2. Similar procedures for generating seed HMMs with $\Sigma$ perturbation and transition matrix perturbation are presented in Sec. 3 of the supplementary material. Briefly speaking, 1) for $\Sigma$ perturbation, a perturbed $\Sigma$ is generated by linear combination of a given $\Sigma_{base}$ and a random matrix $\Sigma_{pert}$ drawn from a Wishard distribution; 2) for transition matrix perturbation, a perturbed transition matrix is generated by linear combination of a given $\mathbf{T}_{base}$ and a random matrix $\mathbf{P}$, each row of which is drawn from a Dirichlet distribution. In both cases, the magnitude of perturbation can be controlled by the coefficients in the linear combinations.

We explored a range of hyper-parameters for the HMMs. In particular, we let the number of states be 3, 5, or 8, the

TABLE 1: Summary of the parameters setup for parameter perturbation experiments. $rand(2)$ here means random matrix of dimension 2 by 2. $Dirichlet(\vec{x})$ here means generating samples from Direchlet distribution with parameter $\vec{x}$.

| Exp. index | deviation step | $\vec{\mu}$ | $\vec{\Sigma}$ | $\mathbf{T}$ |
|---|---|---|---|---|
| 1 | $\Delta\mu = 0.2$, 0.4, 0.6 | $\left\{ \begin{pmatrix} 2+i\Delta\mu \\ 2+i\Delta\mu \end{pmatrix}, \begin{pmatrix} 5+i\Delta\mu \\ 5+i\Delta\mu \end{pmatrix} \right.$ $\left. \vert i = 1,2,3,4,5 \right\}$ | $\left\{ \begin{pmatrix} 1 & 0 \\ 0 & 1 \end{pmatrix}, \begin{pmatrix} 1 & 0 \\ 0 & 1 \end{pmatrix} \right\}$ | $\begin{pmatrix} 0.8 & 0.2 \\ 0.2 & 0.8 \end{pmatrix}$ |
| 2 | $\Delta\sigma = 0.2$, 0.4, 0.6 | $\left\{ \begin{pmatrix} 2 \\ 2 \end{pmatrix}, \begin{pmatrix} 5 \\ 5 \end{pmatrix} \right\}$ | $\{\{0.2 \cdot exp(i\Delta\sigma \cdot \mathbf{S}),$ $0.2 \cdot exp(i\Delta\sigma \cdot \mathbf{S})\} \vert$ $i = 1,2,3,4,5\},$ $\mathbf{S} = rand(2)$ | $\begin{pmatrix} 0.8 & 0.2 \\ 0.2 & 0.8 \end{pmatrix}$ |
| 3 | $\Delta t = 0.2$, 0.4, 0.6 | $\left\{ \begin{pmatrix} 2 \\ 2 \end{pmatrix}, \begin{pmatrix} 5 \\ 5 \end{pmatrix} \right\}$ | $\left\{ \begin{pmatrix} 1 & 0 \\ 0 & 1 \end{pmatrix}, \begin{pmatrix} 1 & 0 \\ 0 & 1 \end{pmatrix} \right\}$ | $\{\Delta t \cdot \mathbf{S} + (1-\Delta t) \cdot \mathbf{T}_i \vert$ $\mathbf{T}_i[j,:] \sim Dirichlet(10 \cdot \mathbf{S}[j,:])$ $i = 1,2,3,4,5\}, \mathbf{S} = \begin{pmatrix} 0.8 & 0.2 \\ 0.2 & 0.8 \end{pmatrix}$ |

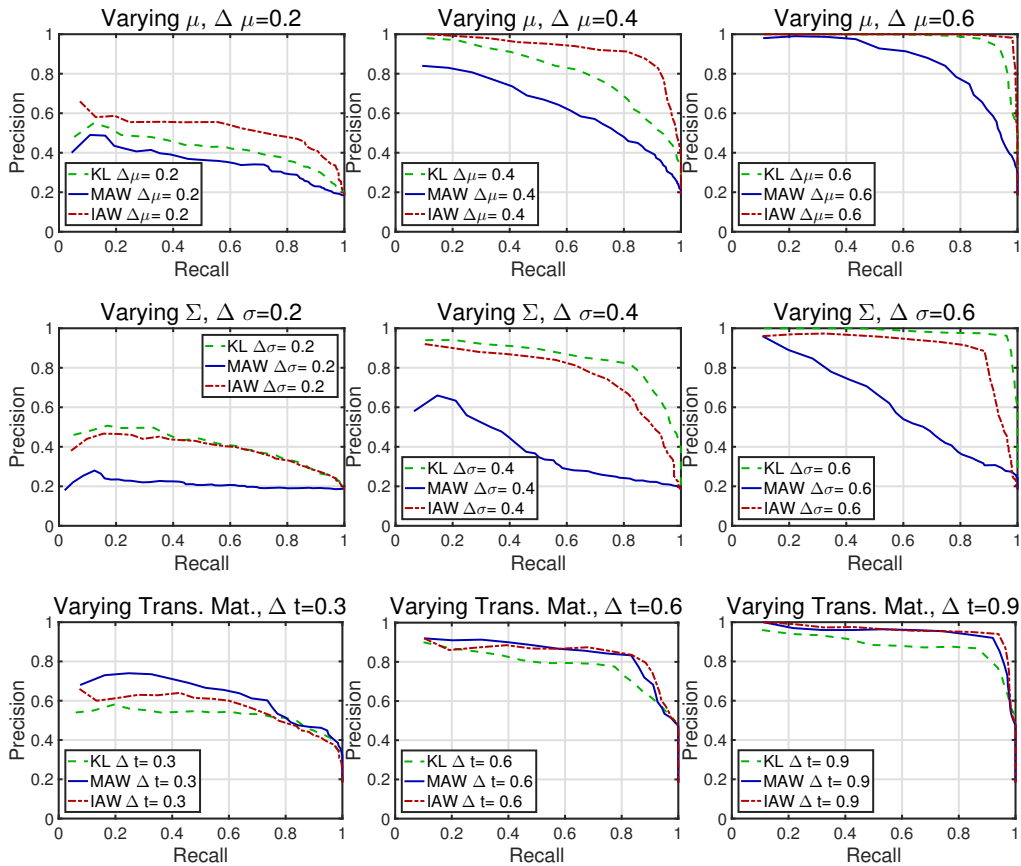

Fig. 3: Precision-recall plot for the study to compare KL, MAW and IAW's sensitivity to the perturbation of GMM-HMM's parameters.

TABLE 2: Synthetic data per distance computation time comparison. ( KL sample size: 2000, IAW sample size: 500)

| | KL | MAW | IAW |
|---|---|---|---|
| time | 5.8ms | 3.8ms | 57.9ms |

dimension be 3, 5, or 8, and set the extent of differentiation between HMMs to 4 levels. Any combination of choices is experimented with, resulting in performance evaluations for $3 \times 3 \times 4 = 36$ data sets. The number of seed HMMs in each data set, denoted by $S$, is 6; and the number of sequences

(and hence estimated HMMs) generated from each seed HMM is set to 6.

The experimental results are displayed in Figure 4. For each case of perturbation, results are shown in a panel of 3 by 3 subfigures. The dimension of HMMs is in increasing order from top rows to bottom rows, and the number of states is in increasing order from left columns to right columns. Within each subfigure, the dimension and the number of states for HMMs are fixed, while the horizontal axis is for the value of $\gamma$, the so-called "scale", by which we control the extent of differentiation between the HMMs. The vertical axis is for value of Area Under the receiver

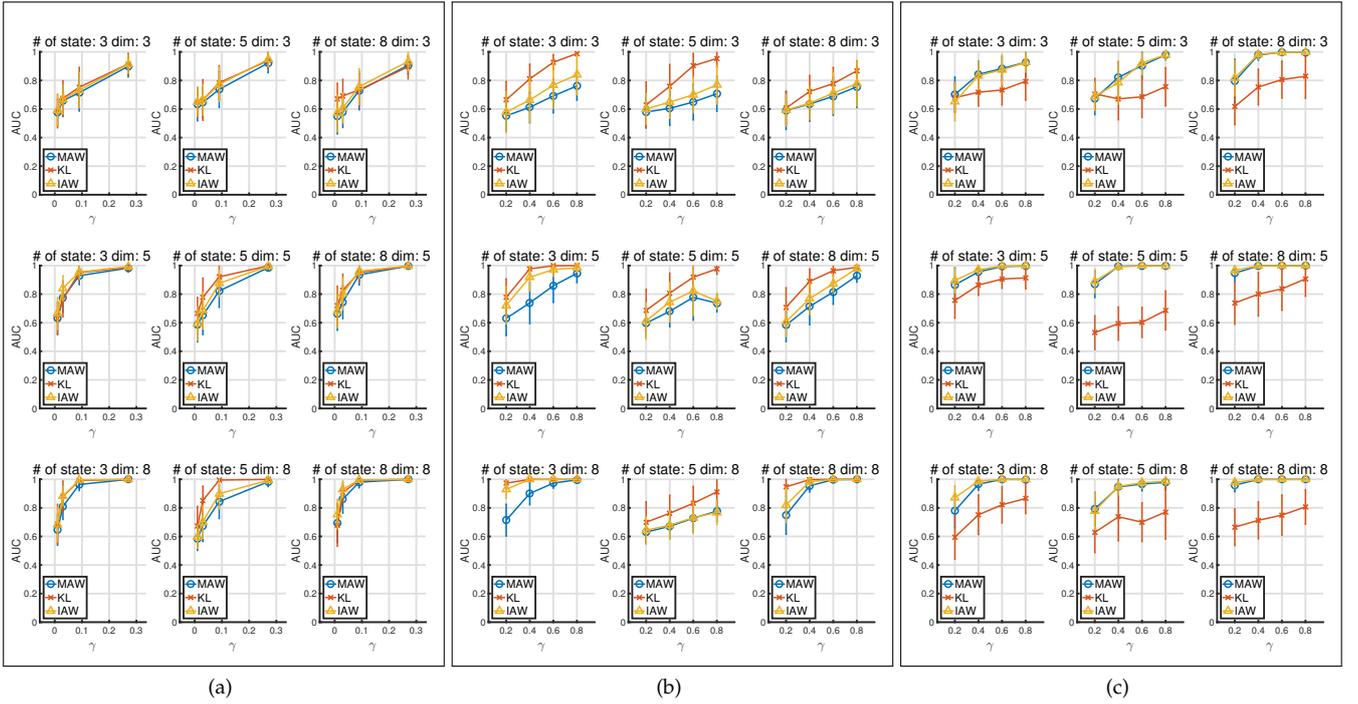

Fig. 4: NN Retrieval experiment Scale v.s. AUCs. under (a) perturbation of $\mu$, (b) perturbation of $\Sigma$ and (c) perturbation of transition matrix.

---

**Algorithm 2** Seed HMMs generation for $\mu$ perturbation experiment.

**Input:** State number: $M$, dimension: $D$, transition matrix $\mathbf{T} \in \mathbb{R}^{M \times M}$, number of seed HMMs: $S$. perturbation magnitude control variable: $\gamma$

**Output:** $\{\Lambda_i\left(\mathbf{T}, \mathcal{M}\left(\{\mu_{j,i}\}_{j=1}^{M}, \vec{\Sigma}\right)\right)\}_{i=1}^{S}$

1: Draw $M$ seed $\mu$s, i.e. $\{\mu_1, ..., \mu_M\}$, from $\mathcal{N}(\vec{0}, \mathbf{I})$, .
2: For each seed $\mu_i$ $i \in 1, ..., M$, we sample $S$ $\mu$s, i.e. $\{\mu_{i,1}, \mu_{i,2}, ..., \mu_{i,S}\}$ from $\mathcal{N}(\mu_i, \gamma \cdot \mathbf{I})$ as the $i$-th state of each HMM's $\mu$ parameters. where $\gamma$ is a knob to control the magnitude of the perturbation.
3: Set Every seed HMM the same $\vec{\Sigma} = \{\Sigma_1, ..., \Sigma_M\}$, $\forall i \in \{1, ..., M\}$, $\Sigma_i$ is a $D \times D$ matrix $\beta_i \times \mathbf{I}$, where $\beta_i \sim \mathcal{N}(1, 0.1)$.
4: Set Every seed HMM the same transition matrix $\mathbf{T}$.

---

operating characteristic Curve (AUC), which is a widely used evaluation metric for retrieval. The higher the value, the better the retrieval performance. Since we use every generated HMM as a query in each data set, we have obtained an error bar computed from all the queries under any setting, which is also shown in the subfigures.

In general, We can see from Figure 4 that the conclusion we made in Section 5 holds across different numbers of states, dimensions, and levels of differentiation between HMMs. More specifically,

1) For $\mu$ perturbation experiments, IAW and KL perform similarly, both better than MAW across different numbers of states and dimensions.

2) For $\Sigma$ perturbation experiments, KL performs better than IAW and IAW performs better than MAW.
3) And for transition matrix perturbation experiments, MAW and IAW perform similarly and they both perform much better than KL.

### 5.3 Real Data: Motion Time Series

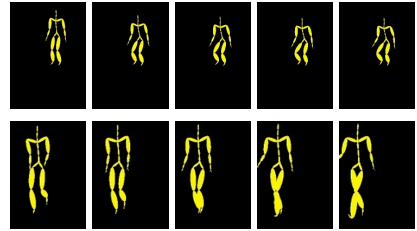

Fig. 5: Visualization of CMU motion capture data. Top: Jump. Bottom: Walk

In this section, we use Carnegie Mellon Motion Capture Dataset (Mocap) to evaluate MAW and IAW and make comparison with KL based approach, which [20] takes. To improve the stability of evaluation, we only select motion categories 1) whose sequences contain only 1 motion, and 2) which contain more than 20 sequences. In total, there are 7 motion categories, i.e. *Alaskan vacation*, *Jump*, *Story*, *clean*, *salsa dance*, *walk*, and *walk on uneven terrain* that meet this criterion and they contain a total of 337 motion sequences. Since the sequence data is of high dimension (62), following the practice of [20], we split the 62 dimension data to 6 joint-groups [1]. And we conduct both motion retrieval based on every joint-group separately and motion classification using Adaboost based on all the joint-groups together.

### 5.3.1 Motion Retrieval

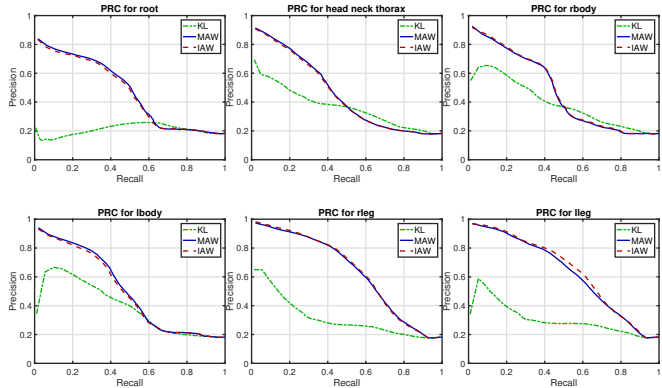

Fig. 6: Precision Recall Plot for Motion Retrieval. The plot for 6 joint-groups, i.e. $root_{12}$, $head\_neck\_thorax_{12}$, $rbody_{12}$, $lbody_{12}$, $rleg_6$, $lleg_6$, are displayed separately.

For each motion time series, we first estimate a 3-state GMM-HMM for each joint-group. Given any joint-group, the model estimated from every sequence is used as query to retrieve models estimated from other sequences using KL, MAW and IAW distances respectively. The parameter $\alpha$ for MAW and IAW is chosen such that the 1-nearest neighbor classification accuracy on a small set-aside evaluation set is maximized. In the supplementary material Sec. 2, an example of how $\alpha$ for $rleg$ is chosen is illustrated in Fig. 1 and the values of all chosen $\alpha$s are displayed in Table 1. The precision-recall plot for the motion retrieval is shown in Fig. 6. Any point on the precision-recall curve is an average over all the motion sequences, each served as query once. We can see that MAW and IAW yield consistently better retrieval results on all the joint-groups.

### 5.3.2 Retrieval with Missing Dimensions

One important advantage of MAW or IAW over KL is that it can handle HMMs with degenerated support. Imagine a scenario that a proportion of collected time series may have missing dimensions for various reasons such as malfunctioning sensor(s). This situation can arise frequently in real world data collection when sensor monitoring expands over a long period in a highly dynamic environment. If we are restricted to use complete data, it is possible that only a small fraction of instances qualify.

For simplicity of experiments, we still conduct retrieval experiments using the motion time series and simulate the scenario when 1, 2 or 3 sensors malfunctioned so we failed to record the data for the corresponding dimensions. Specifically, each motion sequence is used as query once. For any query motion sequence, we randomly select 1, 2 or 3 dimensions to assume missing and set the values of those dimensions to 0.0. As described in Section 5.3, we retrieve sequences based on the MAW distance between the models estimated from the sequences. The model estimated from the query is a degenerated distribution with the variances of the missing dimensions being zero. The models estimated from the other sequences are based on complete data. The precision recall curves are plotted in Fig.7. The performance drops as the number of missing dimensions increases. However, the degradation is small in comparison with the difference between MAW and KL. If we compare Fig. 7 and 6, we see that even when every query has 3 missing dimensions, the most part of the precision-recall curve is clearly better than that achieved by KL in every subfigure (each corresponding to one joint-group). Similar experiments using the IAW distance have also been done with results shown in the supplementary material Sec. 4. We can draw the same conclusion for IAW.

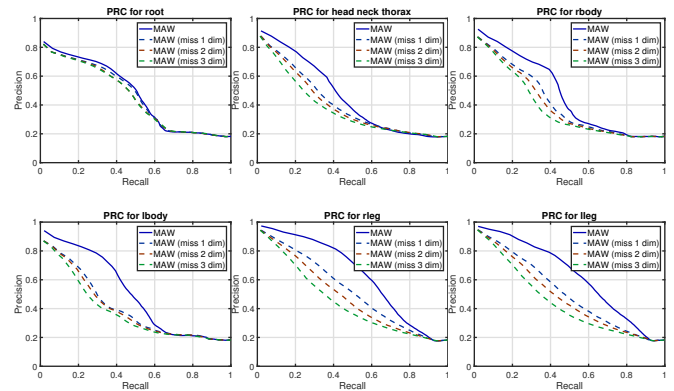

Fig. 7: Precision Recall Plot for MAW based Motion Retrieval under data missing for certain dimension(s) setting. The plot for 6 joint-groups, i.e. $root_{12}$, $head\_neck\_thorax_{12}$, $rbody_{12}$, $lbody_{12}$, $rleg_{12}$, $lleg_6$, are displayed separately.

### 5.3.3 Motion Classification

First, we split the 337 motion sequences randomly into two sets, roughly half for training and half for testing. In the training phase, for each of the 7 motion categories, we train one GMM-HMM for every joint-group. For each sequence, we also estimate one GMM-HMM for every joint-group. We then compute its distances (either KL, MAW or IAW) to all the 7 GMM-HMMs (one for each motion category) on the same joint-group data. We repeat this for every joint-group. These distance values are treated as features. The dimension of the feature vector of one motion sequence is thus the number of joint-groups multiplied by 7. Finally, we use Adaboost with depth-two decision trees to obtain a classification accuracy on the test data. We plot the classification accuracy with respect to the iteration number in Adaboost in Fig. 8 (a). According to the mapping between dimension indexes and sensor locations on the body, the variables of Mocap data can also be split into 27 more refined joint-groups. Under the 27 joint-group split scheme, we run the same classification experiments again and summarize the results in Fig. 8 (b). Overall, the results show that under both the 6 joint-group scheme and the 27 joint-group scheme, MAW (92.90% for 6 joint scheme and 94.67% for 27 joint

---

1. $root_{12}$, $head\_neck\_thorax_{12}$, $rbody_{12}$, $lbody_{12}$, $rleg_6$, $lleg_6$. (The subscript number denotes the dimension of the group) scheme) and IAW (93.49% for 6 joint scheme and 98.22% for 27 joint scheme) achieve considerably better classification accuracy than KL (90.53% for 6 joint scheme and 95.86% for 27 joint scheme ). The confusion matrices are also drawn in the supplementary material Sec. 6.

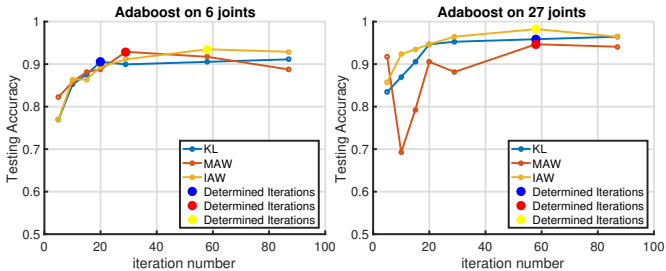

Fig. 8: Testing accuracies with respect to the iteration number in Adaboost (number of weak classifiers selected). (a) Motion Classification by Adaboost on 6 joints. (b) Motion Classification by Adaboost on 27 joints. The iteration number means the number of features incrementally acquired in Adaboost.

The computation time of Mocap data with 6 joint-groups is 21ms by MAW, 158ms by IAW (1000 samples), and 8ms by KL (1000 samples). And that of Mocap data with 27 joint-groups is 17ms by MAW, 160ms by IAW (1000 samples), and 7ms by KL (1000 samples). Again, the MAW and IAW are implemented in MATLAB, and KL-D is implemented in C.

### 5.4 Real Data: TIMIT speech data

The TIMIT[2] data set contains 6300 spoken utterances, each of which is segmented based on 61 phonemes. Following the standard regrouping of phoneme labels described in [35], we select 48 phonemes from 61 phonemes for modeling and these 48 phonemes are merged into 39 phoneme categories. For the stability of HMM estimation, we (randomly) group phoneme segments within each phoneme category into a set of subgroups, each containing 20 phoneme. We also force all phoneme segments within each subgroup either belonging to training set or test set. We concatenate phonemes within each subgroup (Obviously, they have the same phoneme label) and treat it as a single instance for further visualization and classification task. We call such concatenated sequences 20-*concat phoneme segments*. In the training set, we have 2218 such 20-*concat phoneme segments* and in the test set, we have 795 such 20-*concat phoneme segments*. The histogram for the number of 20-*concat phoneme segments* in each phoneme category is shown in the supplementary material Sec. 5.

#### 5.4.1 Phoneme k-NN classification.

We first conduct phoneme classification experiments on TIMIT database using $k$-nearest neighbor based on MAW, IAW and KL distances. The split into training and testing data is typical in the literature (see [36]). As mentioned previously, we use 20-*concat phoneme segments* as instances for classification. We preprocessed mel-frequency cepstral

---

TABLE 3: Phoneme k-NN classification accuracy comparison.

| $k$ | KL | MAW | IAW |
| --- | --- | --- | --- |
| 1 | 0.492 | 0.525 | 0.532 |
| 2 | 0.487 | 0.502 | 0.535 |
| 3 | 0.525 | 0.536 | 0.560 |
| 4 | 0.522 | 0.536 | 0.566 |
| 5 | 0.548 | **0.553** | **0.584** |
| 6 | 0.537 | 0.545 | 0.570 |
| 7 | 0.545 | 0.553 | 0.579 |
| 8 | 0.545 | 0.553 | 0.570 |
| 9 | 0.558 | 0.551 | 0.566 |
| 10 | **0.560** | 0.547 | 0.571 |
| 11 | 0.550 | 0.552 | 0.570 |
| 12 | 0.551 | 0.547 | 0.567 |

coefficients (MFCC features) with sliding window size 4 ms and frame rate 2 ms (and $\Delta$ and $\Delta\Delta$ of MFCC). Following [37], we estimate an HMM with 3 states for each segmented phoneme which is a sequence of 39-dimensional MFCC feature vectors (13 static coefficients, $\Delta$, and $\Delta\Delta$). Following common practices in speech research [38], we force the $\Sigma$'s of HMM to be diagonal. For each HMM estimated from a 20-*concat phoneme segment* in the test set, we compute its KL, MAW and IAW distances to all the HMMs estimated from 20-*concat phoneme segments* in the training set. For MAW and IAW, $\alpha$ is selected such that 1-nearest neighbor accuracy is maximized on the training set. We set the sample size for KL to 5000 and sample size for IAW to 500. Then we perform $k$-nearest neighbor classification. The accuracies with respect to $k$ are shown in Table 3. The best accuracy for KL, MAW or IAW respectively is underscored. IAW achieves significantly better accuracy than MAW and KL, while MAW and KL yield similar results.

#### 5.4.2 Phoneme t-SNE visualization.

We also apply t-SNE visualization[39] to all the 20-*concat phoneme segments* (including both training set and test set). The t-SNE visualization method only relies on a pair-wise distance matrix **D** for all the instances. So we first use KL, MAW and IAW to compute $\mathbf{D}_{KL}$, $\mathbf{D}_{MAW}$, $\mathbf{D}_{IAW}$ respectively, then feed each of them to the t-SNE method, and finally compare the visualization results in Figure 9. We argue that t-SNE based on MAW or IAM shows better visualization than that based on KL because the clusters appear more distinct and compact. We want to emphasize that we use HMMs estimated from 20-*concat phoneme segments* as visualization instances, whereas [39] uses raw MFCC features of each phoneme segment *frame* as a visualization instance (specifically, **D** is obtained from Euclidean distances between per frame features, all of which are of the same size). In short, our visualization for the HMMs takes into account both the difference in the marginal distributions of the feature vectors and the difference in stochastic transition, whereas Figure 7 in [39] only captures the first kind of difference.

## 6 DISCUSSION AND CONCLUSIONS

Although we focus on GMM-HMM whose emission function is Gaussian in this paper, the same methodology extends readily to :

---

2. https://catalog.ldc.upenn.edu/ldc93s1

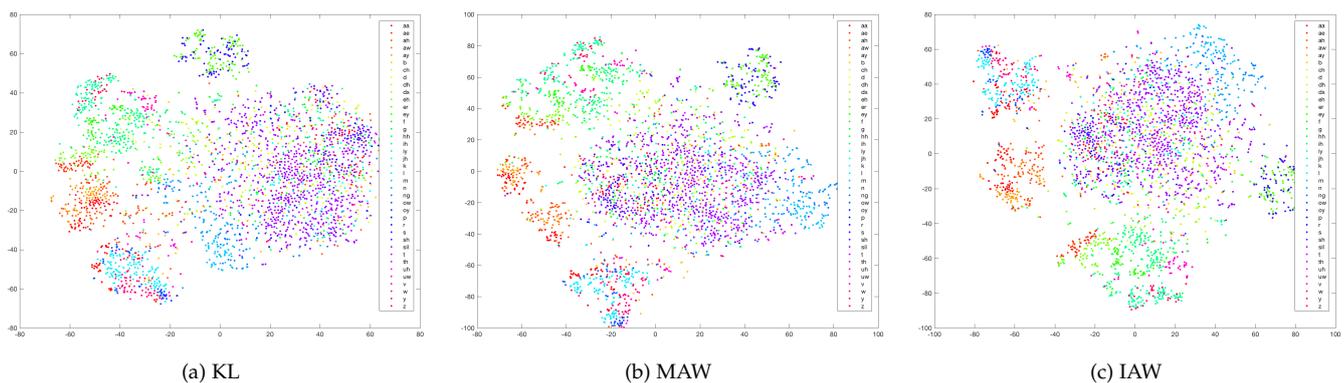

(a) KL  (b) MAW  (c) IAW

Fig. 9: KL-based, MAW-based and IAW-based t-SNE visualization comparison.

1) GMM-HMM whose emission function is GMM but *not* single Gaussian. (Each state with GMM emission function consists of $k$ Gaussians can be split into $k$ states. Our current method can be applied directly then.)

2) Other Hidden Markov Models with non-Gaussian state emission functions, provided that a distance between any two state conditional distributions can be computed. For instance, an HMM with discrete emission distributions can be handled by using the Wasserstein metric between discrete distributions.

In conclusion, we have developed the MAW and IAW distances between GMM-HMMs that are invariant to state permutation. These new distances are computationally efficient, especially MAW. Comparisons with the KL divergence have demonstrated stronger retrieval and classification performance and improved t-SNE visualization. In the future, it is interesting to explore how to reasonably group HMMs into a number of clusters based on our proposed MAW and IAW. The HMM clustering has been studied under the context of KL divergence [13], and the clustering under Wasserstein distance has been studied for empirical distributions [40].

### ACKNOWLEDGMENTS

This research is supported by the National Science Foundation under grant number ECCS-1462230.

### DETAILS OF THE CODE

The code for this work can be found at: https://github.com/cykustcc/aggregated_wasserstein_hmm

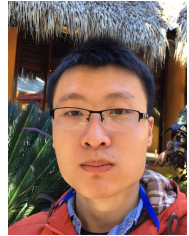

**Yukun Chen** received his BS degree in Applied Physics from the University of Science and Technology of China in 2014. He is currently a PhD candidate and Research Assistant at the College of Information Sciences and Technology, The Pennsylvania State University. He has been a software engineering intern at Google in 2017. His research interests include statistical modeling and learning, computer vision and data mining.

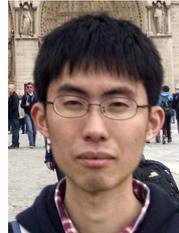

**Jianbo Ye** received his BS degree in Mathematics from the University of Science and Technology of China in 2011. He worked as a research postgraduate at The University of Hong Kong, from 2011 to 2012, and as a research intern at Intel Labs China in 2013 and Adobe in 2017. He is currently a PhD candidate and Research Assistant at the College of Information Sciences and Technology, The Pennsylvania State University. His research interests include statistical modeling and learning, numerical optimization, and human computation for interdisciplinary studies.

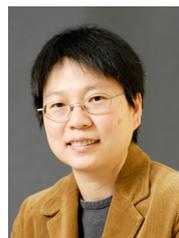

**Jia Li** is a Professor of Statistics at The Pennsylvania State University. She received the MS degree in Electrical Engineering, the MS degree in Statistics, and the PhD degree in Electrical Engineering, all from Stanford University. She worked as a Program Director in the Division of Mathematical Sciences at the National Science Foundation from 2011 to 2013, a Visiting Scientist at Google Labs in Pittsburgh from 2007 to 2008, a researcher at the Xerox Palo Alto Research Center from 1999 to 2000, and a Research Associate in the Computer Science Department at Stanford University in 1999. Her research interests include statistical modeling and learning, data mining, computational biology, image processing, and image annotation and retrieval.

# APPENDIX A
## PROOF OF THEOREM 2

*Proof 3.* Since Wasserstein Distance is a metric[3],

$$\widetilde{R}_p(\mathcal{M}_1, \mathcal{M}_2; \mathbf{W}) \geq 0, \tag{23a}$$

$$\widetilde{R}_p(\mathcal{M}_1, \mathcal{M}_2; \mathbf{W}) = \widetilde{R}_2(\mathcal{M}_2, \mathcal{M}_1; \mathbf{W}), \tag{23b}$$

$$d_T(\mathbf{T}_1, \widetilde{\mathbf{T}}_2)^p = \sum_{i=1}^{M_1} \pi_{1,i} \widetilde{W}_2 \left( \mathcal{M}_1^{(i)}|_{\mathbf{T}_1(i,:)}, \mathcal{M}_1^{(i)}|_{\widetilde{\mathbf{T}}_2(i,:)} \right)^p$$
$$\geq 0, \tag{24a}$$

$$d_T(\mathbf{T}_2, \widetilde{\mathbf{T}}_1)^p = \sum_{i=1}^{M_2} \pi_{2,i} \widetilde{W}_2 \left( \mathcal{M}_2^{(i)}|_{\mathbf{T}_2(i,:)}, \mathcal{M}_2^{(i)}|_{\widetilde{\mathbf{T}}_1(i,:)} \right)^p$$
$$\geq 0. \tag{24b}$$

By Eq. (23a), (24a) and (24b),

$$MAW(\Lambda_1, \Lambda_2) \tag{25}$$

$$\overset{\text{def}}{=} (1-\alpha) \widetilde{R}_p(\mathcal{M}_1, \mathcal{M}_2; \mathbf{W}) + \alpha D_p(\mathbf{T}_1, \mathbf{T}_2 : \mathbf{W}) \tag{26}$$

$$\geq 0. \tag{27}$$

And

$$D_p(\mathbf{T}_1, \mathbf{T}_2 : \mathbf{W})^p = d_T(\mathbf{T}_1, \widetilde{\mathbf{T}}_2)^p + d_T(\mathbf{T}_2, \widetilde{\mathbf{T}}_1)^p$$
$$= d_T(\mathbf{T}_2, \widetilde{\mathbf{T}}_1)^p + d_T(\mathbf{T}_1, \widetilde{\mathbf{T}}_2)^p$$
$$= D(\mathbf{T}_2, \mathbf{T}_1 : \mathbf{W})^p. \tag{28}$$

By, Eq. (23b), (28),

$$MAW(\Lambda_1, \Lambda_2) = MAW(\Lambda_2, \Lambda_1). \tag{29}$$

So we have proved MAW is symmetric, greater or equal than zero. And it's obvious that MAW has zero distance between two GMM-HMMs who represent the same distribution. The remaining part is to prove if two GMM-HMMs have zero MAW distance, their distributions are the same.

If $MAW(\Lambda_1, \Lambda_2) = 0$, because $0 < \alpha < 1$ and by Eq. (23a),(24a) and (24b),

$$\widetilde{R}_p(\mathcal{M}_1, \mathcal{M}_2; \mathbf{W}) = 0 \tag{30a}$$

$$D_p(\mathbf{T}_1, \mathbf{T}_2 : \mathbf{W}) = 0 \tag{30b}$$

By Eq. (30a) and the fact that Wasserstein distance for Gaussian is a true metric, $\mathcal{M}_1$ and $\mathcal{M}_2$ should be identical.

By Eq. (24a), (24b), and the fact that Wasserstein distance for Gaussian is a true metric

$$\widetilde{W}_p \left( \mathcal{M}_1^{(i)}|_{\mathbf{T}_1(i,:)}, \mathcal{M}_1^{(i)}|_{\widetilde{\mathbf{T}}_2(i,:)} \right) = 0 \tag{31}$$

$$\widetilde{W}_p \left( \mathcal{M}_2^{(i)}|_{\mathbf{T}_2(i,:)}, \mathcal{M}_2^{(i)}|_{\widetilde{\mathbf{T}}_1(i,:)} \right) = 0 \tag{32}$$

That is $\mathcal{M}_1^{(i)}|_{\mathbf{T}_1(i,:)}$, $\mathcal{M}_1^{(i)}|_{\widetilde{\mathbf{T}}_2(i,:)}$ are identical and $\mathcal{M}_2^{(i)}|_{\mathbf{T}_2(i,:)}$, $\mathcal{M}_2^{(i)}|_{\widetilde{\mathbf{T}}_1(i,:)}$ are identical. So $\mathbf{T}_1 = \mathbf{T}_2$. Then $\Lambda_1(\mathcal{M}_1, \mathbf{T}_1)$ and $\Lambda_2(\mathcal{M}_2, \mathbf{T}_2)$ should be identical. So, we have proved $MAW$ is a semi-metric.

Note that by Eq. (23a), (23b) and the fact that $\widetilde{R}_2(\mathcal{M}_1, \mathcal{M}_2; \mathbf{W}) = 0$ iff $\mathcal{M}_1$ and $\mathcal{M}_2$ are identical, we also proved that $\widetilde{R}_2(\mathcal{M}_1, \mathcal{M}_2; \mathbf{W})$ is a metric for GMM. (We mentioned this at Section 3.2) ∎

# APPENDIX B
## PROOF OF THEOREM 3

*Proof 4.* For the ease of notation, we assume $p = 2$. The proof also applies to any $0 < p \leq 2$ under trivial modification (implied by Hölder inequality). We let

$$\widehat{\Pi}(\mathcal{M}_1, \mathcal{M}_2) \overset{\text{def}}{=} \left\{ \gamma \overset{\text{def}}{=} \sum_{i=1}^{M_1} \sum_{j=1}^{M_2} w_{i,j} \gamma_{i,j} \right|$$

$$\gamma_{i,j} \in \Pi(\tilde{\phi}_{1,i,j}, \tilde{\phi}_{2,i,j}), \{\tilde{\phi}_{1,i,j}, \tilde{\phi}_{2,i,j}\} \in \Gamma(\mathbf{W}|\mathcal{M}_1, \mathcal{M}_2),$$

$$\mathbf{W} \in \Pi(\pi_1, \pi_2) \bigg\}.$$

From the definitions, one can verify that $\widetilde{\Pi}(\mathcal{M}_1, \mathcal{M}_2) \subseteq \widehat{\Pi}(\mathcal{M}_1, \mathcal{M}_2) \subseteq \Pi(\mathcal{M}_1, \mathcal{M}_2)$. Therefore, optimizing transportation cost over $\gamma \in \widetilde{\Pi}(\mathcal{M}_1, \mathcal{M}_2)$ gives a tighter upper bound of $W(\mathcal{M}_1, \mathcal{M}_2)$ than $\widetilde{W}(\mathcal{M}_1, \mathcal{M}_2)$. Moreover, Theorem 3 is proved if we have

$$\widehat{\Pi}(\mathcal{M}_1, \mathcal{M}_2) = \Pi(\mathcal{M}_1, \mathcal{M}_2).$$

To prove this, we only need to show that for any $\gamma \in \Pi(\mathcal{M}_1, \mathcal{M}_2)$, there exist $w_{i,j} \in \Pi(\pi_1, \pi_2)$, $\{\tilde{\phi}_{1,i,j}, \tilde{\phi}_{2,i,j}\} \in \Gamma(\{w_{i,j}\}|\mathcal{M}_1, \mathcal{M}_2)$ and $\gamma_{i,j} \in \Pi(\tilde{\phi}_{1,i,j}, \tilde{\phi}_{2,i,j})$ with $i = 1, \ldots, M_1$ and $j = 1, \ldots, M_2$ such that

$$\gamma = \sum_{z_1=1}^{M_1} \sum_{z_2=1}^{M_2} w_{z_1,z_2} \gamma_{z_1,z_2}.$$

The constructive proof goes in two steps. First, given any random variables $(x_1, x_2) \sim \gamma \in \Pi(\mathcal{M}_1, \mathcal{M}_2)$, we can induce component membership random variables $(z_1, z_2)$ by

$$p(z_1, z_2) = \int_{\mathbb{R}^d \times \mathbb{R}^d} p(z_1, z_2 | x_1, x_2) d\gamma(x_1, x_2) \overset{\text{def}}{=} w_{z_1, z_2}, \tag{33}$$

where the condition probability is defined multiplicatively by

$$p(z_1, z_2 | x_1, x_2) \overset{\text{def}}{=} \frac{\pi_{1,z_1} \phi_{1,z_1}(x_1)}{f_1(x_1)} \cdot \frac{\pi_{2,z_2} \phi_{2,z_2}(x_2)}{f_2(x_2)}. \tag{34}$$

One can verify that $\{w_{i,j}\} \in \Pi(\pi_1, \pi_2)$ by the definition of Eq. (33): for any $i = 1, \ldots, M_1$

$$\sum_{j=1}^{M_2} w_{i,j} = \int_{\mathbb{R}^d \times \mathbb{R}^d} \sum_{j=1}^{M_2} p(z_1 = i, z_2 = j | x_1, x_2) d\gamma(x_1, x_2)$$

$$= \int_{\mathbb{R}^d \times \mathbb{R}^d} \frac{\pi_{1,i} \phi_{1,i}(x_1)}{f_1(x_1)} d\gamma(x_1, x_2)$$

$$= \int_{\mathbb{R}^d} \pi_{1,i} \phi_{1,i}(x_1) dx_1$$

(integral out $x_2$, since $\gamma \in \Pi(\mathcal{M}_1, \mathcal{M}_2)$)

$$= \pi_{1,i}.$$

Likewise, $\sum_{i=1}^{M_1} w_{i,j} = \pi_{2,j}$ for any $j = 1, \ldots, M_2$. It is obvious that $\mathbf{W}$ defined by Eqs. (33) and (34) is the same as defined by Eq. (20).

Second, consider the conditional measure

$$\gamma(x_1, x_2 | z_1, z_2) = \frac{p(z_1, z_2 | x_1, x_2) \gamma(x_1, x_2)}{w_{z_1, z_2}}$$

(by the Bayes rule), its marginals are

$$d\gamma(x_1 | z_1, z_2) = \frac{1}{w_{z_1, z_2}} \int_{x_2 \in \mathbb{R}^d} p(z_1, z_2 | x_1, x_2) d\gamma(x_1, x_2),$$

$$d\gamma(x_2 | z_1, z_2) = \frac{1}{w_{z_1, z_2}} \int_{x_1 \in \mathbb{R}^d} p(z_1, z_2 | x_1, x_2) d\gamma(x_1, x_2).$$

By definition, we know $\gamma(x_1, x_2 | z_1, z_2) \in \Pi(\gamma(x_1 | z_1, z_2), \gamma(x_2 | z_1, z_2))$. One can validate that $\{\gamma(x_1 | z_1, z_2), \gamma(x_2 | z_1, z_2)\} \in \Gamma(\{w_{i,j}\} | \mathcal{M}_1, \mathcal{M}_2)$: for $z_1 = 1, \ldots, M_1$ and $z_2 = 1, \ldots, M_2$,

$$\begin{aligned}
&\sum_{z_2=1}^{M_2} \frac{w_{z_1, z_2}}{\pi_{1, z_1}} d\gamma(x_1 | z_1, z_2) \\
&= \int_{x_2 \in \mathbb{R}^d} \sum_{z_2=1}^{M_2} \frac{\phi_{1, z_1}(x_1)}{f_1(x_1)} \cdot \frac{\pi_{2, z_2} \phi_{2, z_2}(x_2)}{f_2(x_2)} d\gamma(x_1, x_2) \\
&= \int_{x_2 \in \mathbb{R}^d} \frac{\phi_{1, z_1}(x_1)}{f_1(x_1)} d\gamma(x_1, x_2) \\
&= \phi_{1, z_1}(x_1) dx_1.
\end{aligned}$$

Likewise, we can show

$$\sum_{z_1=1}^{M_1} \frac{w_{z_1, z_2}}{\pi_{2, z_2}} d\gamma(x_2 | z_1, z_2) = \phi_{2, z_2}(x_2) dx_2$$

Let $\tilde{\phi}_{l,i,j}$ be the p.d.f. of $\gamma(x_l | z_1 = i, z_2 = j)$ and $\gamma_{i,j} \overset{\text{def}}{=} \gamma(x_1, x_2 | z_1 = i, z_2 = j)$, we see that $\gamma \in \widehat{\Pi}(\mathcal{M}_1, \mathcal{M}_2)$. Therefore, $\Pi(\mathcal{M}_1, \mathcal{M}_2) \subseteq \widehat{\Pi}(\mathcal{M}_1, \mathcal{M}_2)$. Combined with the fact that $\widehat{\Pi}(\mathcal{M}_1, \mathcal{M}_2) \subseteq \Pi(\mathcal{M}_1, \mathcal{M}_2)$, the proof is complete. ∎

***Remark 6.*** The optimal coupling in Eq. (1) can be factored as a finite mixture model with $M_1 \cdot M_2$ components, whose proportion vector is $\mathbf{W}^*$. $\mathbf{W}^*$ is taken from the minimizer $\gamma^* = \sum_{i,j} w_{i,j}^* \gamma_{i,j}^* \in \widehat{\Pi}(\mathcal{M}_1, \mathcal{M}_2)$ of the following problem

$$\inf_{\gamma \in \widehat{\Pi}(\mathcal{M}_1, \mathcal{M}_2)} \int_{\mathbb{R}^d \times \mathbb{R}^d} \|\mathbf{x} - \mathbf{y}\|^2 d\gamma(\mathbf{x}, \mathbf{y}). \qquad (35)$$

# Supplementary material for the paper "Aggregated Wasserstein Metric and State Registration for Hidden Markov Models"

Yukun Chen, Jianbo Ye, and Jia Li

❖

## 1 PROOF OF REMARK 1

*Proof 1.* Let the solved optimal coupling for $W_p$ and $W_q$ be $\gamma_p$ and $\gamma_q$ respectively. To prove $W_p \leq W_q, \forall p \leq q$, i.e.:

$$\left[ \int \|\mathbf{x} - \mathbf{y}\|^p d\gamma_p \right]^{1/p} \leq \left[ \int \|\mathbf{x} - \mathbf{y}\|^q d\gamma_q \right]^{1/q} \quad (1)$$

Since the optimality of $\gamma_p$ for $W_p$, We can prove the stronger inequality below instead to prove $W_p \leq W_q$:

$$\left[ \int \|\mathbf{x} - \mathbf{y}\|^p d\gamma_q \right]^{1/p} \leq \left[ \int \|\mathbf{x} - \mathbf{y}\|^q d\gamma_q \right]^{1/q} \quad (2)$$

Let $f = \|\mathbf{x} - \mathbf{y}\|^p$,

$$\Leftrightarrow \int f d\gamma_q \leq \left[ \int f^{q/p} d\gamma_q \right]^{p/q} \quad (3)$$

Let $\mu = q/p \geq 1$,

$$\Leftrightarrow \int f d\gamma_q \leq \left[ \int f^\mu d\gamma_q \right]^{1/\mu} \quad (4)$$

This holds because:

$$\Leftrightarrow \int f d\gamma_q \leq \left[ \int f^\mu d\gamma_q \right]^{1/\mu} \left[ \int 1^{\mu/(\mu-1)} d\gamma_q \right]^{1-1/\mu} \quad (5)$$

which is exactly the holder inequality. ∎

## 2 CHOOSING $\alpha$

An example of how $\alpha$ for rleg is chosen is illustrated in Fig. 1 and the values of all chosen s are displayed in Table 1.

TABLE 1: Choice of $\alpha$ for MAW and IAW in Motion Retrieval Experiments.

| Distance | root | head neck | rbody | lbody | rleg | lleg |
|---|---|---|---|---|---|---|
| MAW | 0.03 | 0.01 | 0.25 | 0.07 | 0.13 | 0.21 |
| IAW | 0.15 | 0.05 | 0.16 | 0.08 | 0 | 0.08 |

## 3 SEED HMMS GENERATION FOR $\Sigma$ PERTURBATION EXPERIMENTS AND TRANSITION MATRIX PERTURBATION

Seed HMMs generation procedures for $\Sigma$ perturbation experiments and transition matrix perturbation experiments are specified in Algorithm 1 and Algorithm 2 respectively.

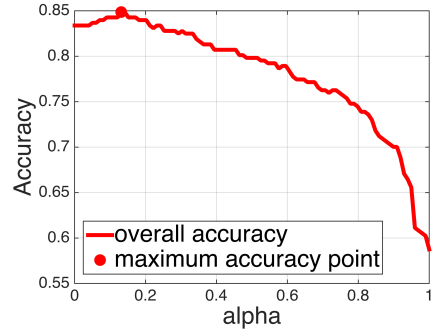

Fig. 1: overall accuracy for *rleg*. w.r.t MAW's $\alpha$

---

**Algorithm 1** Seed HMMs generation for $\Sigma$ perturbation experiment.

**Input:** State number: $M$, dimension: $D$, transition matrix $\mathbf{T} \in \mathbb{R}^{M \times M}$, number of seed HMMs: $S$. base sigma $\Sigma_{base}$, perturbation magnitude control variable: $\gamma$.
**Output:** $\{\Lambda_i (\mathbf{T}, \mathcal{M}(\vec{\mu}, \{\Sigma_{1,i} \dots \Sigma_{M,i}\}))\}_{i=1}^S$
1: Set every seed HMM the same $\vec{\mu} = \{\mu_1, ..., \mu_M\}$, $\forall i \in \{1, ..., M\}, \mu_i \sim \mathcal{N}(\mathbf{0}, 5 \cdot \mathbf{I})$
2: randomly generate $\{\Sigma_i\}_{i=1}^S$s as $(1 - \gamma) \cdot \Sigma_{base} + \gamma \Sigma_i^{pert}$ where $\Sigma_i^{pert} \sim \mathscr{W}_D(\Sigma_{base}, df)$. ($\mathscr{W}_D$ is the Wishard distribution. We set $df$ to 10 in our experiments)
3: randomly generate $\{\beta_j\}_{j=1}^M$ where $\beta_j \sim \mathcal{N}(1, 0.1)$.
4: Set $\{\Sigma'_{j,i} | \Sigma'_{j,i} = \beta_j \Sigma_i, i = 1 \dots S, j = 1 \dots M\}$
5: Set Every seed HMM the same transition matrix $\mathbf{T}$.

---

## 4 MOTION TIME SERIES RETRIEVAL WITH IAW WHEN CERTAIN DIMENSION HAS MISSING DATA.

The results of retrieval with missing dimensions using IAW are shown in Figure 2

## 5 HISTOGRAM OF 20-CONCAT PHONEME SEGMENTS

Histogram of 20-concat phoneme segments used in Section 5.4 is plotted in Fig. 3

---

**Algorithm 2** Seed HMMs generation for transition matrix perturbation experiment.

---

**Input:** State number: $M$, dimension: $D$, base transition matrix $\mathbf{T}_{base} \in \mathbb{R}^{M \times M}$, number of seed HMMs: $S$. perturbation magnitude control variable: $\gamma$.

**Output:** $\left\{ \Lambda_i \left( \mathbf{T}_i, \mathcal{M} \left( \{\mu_j\}_{j=1}^{M}, \{\Sigma_{i,j} | \Sigma_{i,j} = \Sigma_i\}_{j=1}^{M} \right) \right) \right\}_{i=1}^{S}$

1: Set every seed HMM the same $\vec{\mu} = \{\mu_1, ..., \mu_M\}$, $\forall i \in \{1, ..., M\}$, $\mu_i \sim \mathcal{N}(\mathbf{0}, 5 \cdot \mathbf{I})$

2: Set every seed HMM the same $\vec{\Sigma} = \{\Sigma_1, ..., \Sigma_M\}$, $\forall i \in \{1, ..., M\}$, $\Sigma_i$ is a $D \times D$ matrix $\beta_i \mathbf{I}$, where $\beta_i \sim \mathcal{N}(1, 0.1)$.

3: randomly generate transition matrices: $\{\mathbf{T}_i\}_{i=1}^{S}$ where $\mathbf{T}_i = \gamma \cdot \mathbf{T}_{base} + (1-\gamma) \cdot \mathbf{P}_i$, each row of perturbation matrix $\mathbf{P}$ is sampled by: $\mathbf{P}_{i,:} \sim Dir([1, 1, ..., 1]_{1 \times M})$

---

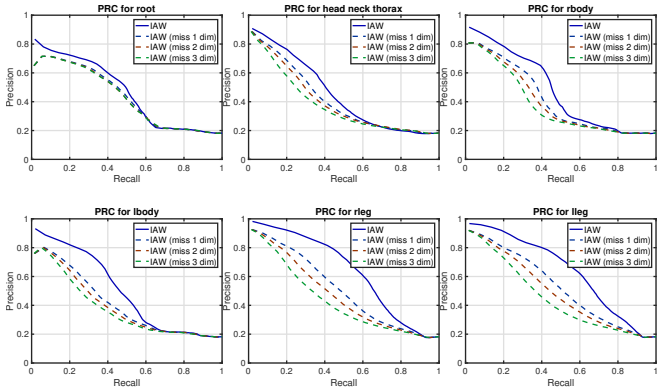

Fig. 2: Precision Recall Plot for IAW based Motion Retrieval under data missing for certain dimension(s) setting. The plot for 6 joint-groups, i.e. $root_{12}$, $head\_neck\_thorax_{12}$, $rbody_{12}$, $lbody_{12}$, $rleg_6$, $lleg_6$, are displayed separately. (Best view in color.)

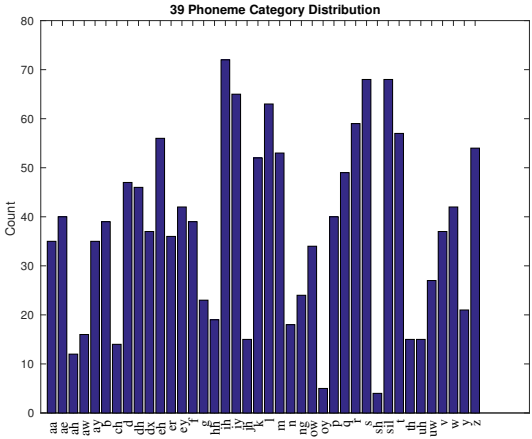

Fig. 3: Histogram of 20-concat phoneme segments

## 6 CONFUSION MATRICES FOR MOTION CLASSIFICATION

Confusion matrices for motion classification experiment in Section 5.3.3 are plotted in Fig. 4

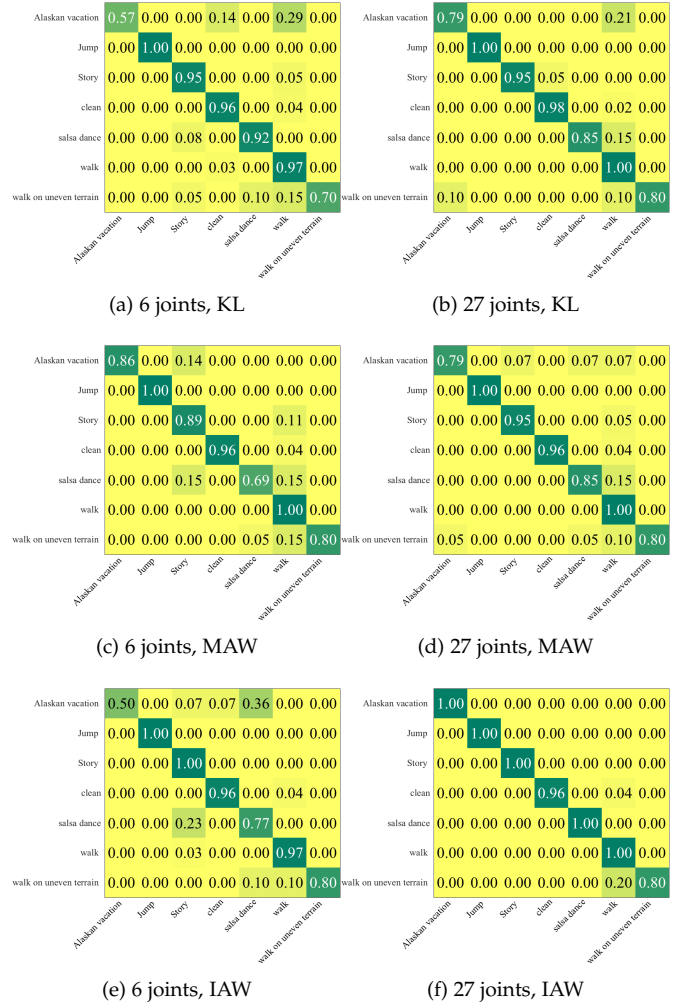

Fig. 4: (a) 6 joints, KL, corresponds to the blue dot in the left subfigure of Fig. 8, (b) 6 joints, MAW, corresponds to the red dot in the right subfigure of Fig. 8, (c) 6 joints, IAW, corresponds to the yellow dot in the left subfigure of Fig. 8, (d) 27 joints, KL, corresponds to the blue dot in the right subfigure of Fig. 8, (e) 27 joints, MAW, corresponds to the red dot in the left subfigure of Fig. 8, (f) 27 joints, IAW, corresponds to the yellow dot in the right subfigure of Fig. 8,